\documentclass[12pt]{article}
\usepackage{fullpage}
\usepackage{graphicx}
\usepackage{epstopdf}
\usepackage{psfrag}
\usepackage{amsmath}
\usepackage{amsfonts}
\usepackage{csquotes}
\usepackage{ifthen}
\usepackage{url}
\usepackage{mathtools}
\usepackage{siunitx}
\usepackage{booktabs}

\newcommand{\ones}{\mathbf 1}
\newcommand{\reals}{{\mbox{\bf R}}}

\newcommand{\symm}{{\mbox{\bf S}}}  

\newcommand{\Tr}{\mathop{\bf Tr}}
\newcommand{\diag}{\mathop{\bf diag}}

\newcommand{\prox}{\mathbf{prox}}

\newcommand{\Prob}{\mathop{\bf Prob}}

\newcommand{\argmin}{\mathop{\rm argmin}}
\newcommand{\argmax}{\mathop{\rm argmax}}

\newcommand{\dom}{\mathop{\bf dom}} 

\newcommand{\sign}{\mathop{\bf sign}}

\newcommand{\eg}{{\it e.g.}}
\newcommand{\ie}{{\it i.e.}}

\newcommand{\BEAS}{\begin{eqnarray*}}
\newcommand{\EEAS}{\end{eqnarray*}}
\newcommand{\BEA}{\begin{eqnarray}}
\newcommand{\EEA}{\end{eqnarray}}
\newcommand{\BEQ}{\begin{equation}}
\newcommand{\EEQ}{\end{equation}}
\newcommand{\BIT}{\begin{itemize}}
\newcommand{\EIT}{\end{itemize}}

\renewcommand{\hat}{\widehat}
\renewcommand{\tilde}{\widetilde}

\newcounter{algorithmctr}[section]
\renewcommand{\thealgorithmctr}{\thesection.\arabic{algorithmctr}}
\newenvironment{algdesc}%
   {\refstepcounter{algorithmctr}\begin{list}{}{%
       \setlength{\rightmargin}{0\linewidth}%
       \setlength{\leftmargin}{.05\linewidth}}%
       \rmfamily\small
       \item[]{\setlength{\parskip}{0ex}\hrulefill\par%
        \nopagebreak{\bfseries\textsf{Algorithm \thealgorithmctr~}}}}%
   {{\setlength{\parskip}{-1ex}\nopagebreak\par\hrulefill} \end{list}}

\title{A Distributed Method for Fitting Laplacian Regularized Stratified Models}
\author{Jonathan Tuck \and Shane Barratt \and Stephen Boyd}

\begin{document}
\maketitle

\begin{abstract}
Stratified models are models that 
depend in an arbitrary way on a set of 
selected categorical features, and depend linearly on
the other features.
In a basic and traditional formulation a separate model is fit
for each value of the categorical feature, using only the
data that has the specific categorical value.
To this formulation we add Laplacian regularization,
which encourages the model parameters for 
neighboring categorical values to be similar.
Laplacian regularization allows us to specify one or more 
weighted graphs on the stratification feature values.
For example, stratifying over the days of the week,
we can specify that the Sunday model parameter should be close to
the Saturday and Monday model parameters.
The regularization improves the performance of the model over
the traditional stratified model, since the model for 
each value of the categorical `borrows strength' from its neighbors.
In particular, it produces a model even for categorical values
that did not appear in the training data set.

We propose an efficient distributed method for fitting stratified models, 
based on the alternating direction method of multipliers (ADMM).
When the fitting loss functions are convex, the stratified model
fitting problem is convex, and our method computes
the global minimizer of the loss plus regularization;
in other cases it computes a local minimizer.
The method is very efficient, and naturally scales to large data
sets or numbers of stratified feature values.
We illustrate our method with a variety of examples.
\end{abstract}

\section{Introduction}
We consider the problem of fitting a model to some given data.
One common and simple paradigm parametrizes the model by a 
parameter vector $\theta$, 
and uses convex optimization to minimize an empirical loss on the 
(training) data set plus a regularization term that (hopefully) skews the 
model parameter toward one for which the model generalizes to new unseen data.
This method usually includes one or more hyper-parameters that scale terms in 
the regularization.
For each of a number of values of the hyper-parameters, a model parameter
is found,  and the resulting model is tested on previously unseen data.  
Among these models, we choose one that achieves a good fit on the test data.
The requirement that the loss function and regularization be convex limits how
the features enter into the model; generally, they are linear (or affine) 
in the parameters.
One advantage of such models is that they are generally interpretable.

Such models are widely used, and often work well in practice.
Least squares regression \cite{legendre1805nouvelles,gauss1809theoria}, 
lasso \cite{tibshirani1996lasso}, logistic regression \cite{cox1958logistic},
support vector classifiers \cite{boser1992training} are common methods that fall
into this category (see \cite{hastie2009elements} for these and others).
At the other extreme, we can use models that in principle fit arbitrary 
nonlinear dependence on the features.  
Popular examples include tree based models \cite{breiman1984classification}
and neural networks \cite{goodfellow2016deeplearningbook}.

\emph{Stratification} is a method to build a model of some data that depends in
an arbitrary way on one or more of the categorical features.
It identifies one or more categorical features and builds a separate model
for the data that take on each of the possible values of these categorical
features.

In this paper we propose augmenting
basic stratification with an additional regularization term.
We take into account the usual loss and regularization terms in the objective,
plus an additional regularization term that encourages the parameters found
for each value to be close to their neighbors on some specified
weighted graph on the categorical values.
We use the simplest possible term that encourages closeness of
neighboring parameter values:
a graph Laplacian on the stratification feature values.
(As we will explain below, several recent papers use more sophisticated
regularization on the parameters for different values of the categorical.)
We refer to a startified model with this objective function as a
\emph{Laplacian regularized stratified model}.
Laplacian regularization can also be interpreted as a
prior that our model parameters 
vary smoothly across the graph on the stratification feature values.

Stratification (without the Laplacian regularization) is an old and simple idea:
Simply fit a different model for each value of the categorical used to stratify.
As a simple example where the data are people, we can stratify on sex, \ie,
fit a separate model for females and males.
(Whether or not this stratified model is better than a common model for both
females and males is determined by out-of-sample or cross-validation.)
To this old idea, we add an additional term that penalizes deviations 
between the model parameters for females and males. (This Laplacian regularization
would be scaled by a hyper-parameter, whose value is chosen by validation on unseen
or test data.)
As a simple extension of the example mentioned above, we can stratify on sex and age,
meaning we have a separate model for females and males, for each age from 1 to 100 (say).
Thus we would create a total of 200 different models: one for each age/sex pair.
The Laplacian regularization that we add encourages, for example, the model parameters 
for 37 year old females to be close to that for 37 year old males, as well as 
36 and 38 year old females.  We would have two hyper-parameters, one that scales
the deviation of the model coefficients across sex, and another that scales 
the deviation of the model coefficients across adjacent ages.
One would choose suitable values of these hyper-parameters by validation on a test
set of data.

There are many other applications where a Laplacian regularized stratified model
offers a simple and interpretable model. We can stratify over time, which gives
time-varying models, with regularization that encourages our time-dependent model
parameters to vary slowly over time.  We can capture multiple periodicities in
time-varying models, for example by asserting that the 11PM model parameter
should be close to the midnight model parameter, the Tuesday and Wednesday parameters 
should be close to each other, and so on.
We can stratify over space, after discretizing location.  The Laplacian regularization
in this case encourages the model parameters to vary smoothly over space.
This idea naturally falls in line with Waldo Tobler's ``first law of geography'' 
\cite{tobler1970computer}:
\begin{displayquote}
Everything is related to everything else, but near things are more related than distant things.
\end{displayquote}

Compared to simple stratified models, our Laplacian regularized stratified models
have several advantages.  Without regularization, stratification is limited
by the number of data points taking each value of the stratification feature,
and of course fails (or does not perform well) when there are no data points with some value
of the stratification feature.
With regularization there is no such limit; we can even build a model when the training 
data has no points with some values of the stratification feature.
In this case the model borrows strength from its neighbors.
Continuing the example mentioned above, we can find a model for 37 year old females,
even when our data set contains none.  Unsurprisingly, the parameter for 37
year old females is a weighted average of the parameters for 36 and 38 year old females,
and 37 year old males.

We can think of stratified models as a hybrid of complex and simple models.
The stratified model has arbitrary dependence on the stratification feature, but simple
(typically linear) dependence on the other features.
It is essentially non-parametric in the stratified feature, and parametric with a 
simple form in the other features.
The Laplacian regularization allows us to come up with a sensible model with 
respect to the stratified feature.

The simplest possible stratified model is a constant for each value 
of the stratification feature. In this case the stratified model is nothing
more than a fully non-parametric model of the stratified feature.
The Laplacian regularization in this case encourages smoothness of the predicted
value relative to some graph on the possible stratification feature values.
More sophisticated models predict a distribution of values of some
features, given the stratified feature value, or build a stratified regression
model, for example.

In this paper, we first describe stratified model fitting with Laplacian 
regularization as a convex optimization problem, with one model parameter vector 
for every value of the stratification feature.  This problem is convex (when the loss 
and regularizers in the fitting method are), and so can be reliably solved \cite{boyd2004convex}.
We propose an efficient distributed solution method based on the 
alternating direction method of multipliers (ADMM) \cite{boyd2011distributed}, 
which allows us to solve the optimization problem in a distributed fashion and 
at very large scale.
In each iteration of our method, we fit the models for each value of the stratified
feature (independently in parallel), and then solve a number of Laplacian systems
(one for each component of the model parameters, also in parallel).
We illustrate the ideas and method with several examples.

\section{Related work}

Stratification, \ie, the idea of separately fitting a different model for 
each value of some parameter, is routinely applied in various practical settings.
For example, in clinical trials, participants are often divided into subgroups, 
and one fits a separate model for the data from each subgroup \cite{kernan1999stratified}.
Stratification by itself has immediate benefits; it can be helpful for
dealing with confounding categorical variables, can help one gain insight into
the nature of the data, and can also play a large role in experiment design 
(see, \eg, \cite{rothman1986epidemiology,kestenbaum2009epidemiology} for some 
applications to clinical research, and \cite{jacob2007efficient} for an application 
to biology).

The idea of adding regularization to fitting stratified models, however, 
is (unfortunately) not as well known in the data science, machine learning, 
and statistics communities.
In this section, we outline some ideas related to stratified models,
and also outline some prior work on Laplacian regularization.

\paragraph{Regularized lasso models.}
The data-shared lasso is a stratified model that encourages closeness of parameters by
their difference as measured by the $\ell_1$-norm \cite{gross2016datasharedlasso}.
(Laplacian regularization penalizes their difference by the $\ell_2$-norm squared.)
The pliable lasso is a generalization of the lasso that is also a stratified model,
encouraging closeness of parameters as measured by a blend of
the $\ell_1$- and $\ell_2$-norms \cite{tibshirani2017pliablelasso,du2018pliablelasso}.
The network lasso is a stratified model that encourages closeness of parameters by
their difference as measured by the $\ell_2$-norm \cite{hallac2015network,hallac2017network}.
The network lasso problem results in groups of models with the
same parameters, in effect clustering the stratification features.
(In contrast, Laplacian regularization leads to smooth parameter values.)
The ideas in the original papers on the network lasso were combined and made more general, 
resulting in the software package SnapVX \cite{hallac2017snapvx}, a general solver for 
convex optimization problems defined on graphs that is based on SNAP \cite{leskovec2016snap}
and cvxpy \cite{diamond2016cvxpy}.
One could, in principle, implement Laplacian regularized stratified models
in SnapVX; instead we develop specialized methods for the specific case of Laplacian
regularization, which leads to faster and more robust methods.
We also have observed that, in most practical settings,
Laplacian regularization is all you need.

\paragraph{Varying-coefficient models.}
Varying-coefficient models are a class of 
regression and classification models in which the model parameters are 
smooth functions of some features \cite{hastie1993varyingcoefmodel}.
A stratified model can be thought of as 
a special case of varying-coefficient models, where the
features used to choose the coefficients are categorical.
Estimating varying-coefficient models is generally nonconvex and 
computationally expensive \cite{fan2008varyingcoefmodel}.

\paragraph{Geographically weighted regression.}
The method of geographically weighted regression (GWR) is widely used
in the field of geographic information systems (GIS).
It consists of a spatially-varying linear model with a location attribute 
\cite{brunsdon1996geographically}.
GWR fits a model for each location using every data point, where
each data point is weighted by a kernel function of the 
pairwise locations.
(See, \eg, \cite{mcmillen2004geographically} for a recent survey and applications.)
GWR is an extremely special case of a stratified model, where the task is regression
and the stratification feature is location.

\paragraph{Graph-based feature regularization.}
One can apply graph regularization to the features themselves.
For example, the group lasso penalty enforces sparsity of model coefficients among
groups of features \cite{YL:06,FHT:10}, performing feature selection at the
level of groups of features.
The group lasso is often fit using a method based on ADMM, which is similar in
vein to the one we describe.
The group lasso penalty can be used as a regularization term to
any linear model, \eg, logistic regression \cite{meier2008group}.
A related idea is to use a Laplacian regularization term on the features,
leading to, \eg, Laplacian regularized least-squares (LapRLS) and Laplacian
regularized support vector machines (LapSVM) \cite{belkin2006manifold}.
Laplacian regularized models have been applied to many disciplines and fields, 
\eg, to semi-supervised learning \cite{zhu2003semi,nadler2009semi_supervised_laplacian}, 
communication \cite{boyd2006laplacianeigenvalues,CWSS:11,ZF:15}, 
medical imaging \cite{ZS:11}, 
computer vision \cite{WZSQS:12}, 
natural language processing \cite{WSZQ:13},
and microbiology \cite{wang2011lapRLSmicrobio}.

\paragraph{Graph interpolation.}
In graph interpolation or regression on a graph, one is given
vectors at some of the vertices of a graph, and is tasked with inferring the
vectors at the other (unknown) vertices \cite{KS:11}.
This can be viewed as a stratified model where the parameter at
each vertex or node is a constant.
This leads to, \eg, the total variation denoising problem in
signal processing \cite{rudin1992nonlinear}.         

\paragraph{Multi-task learning.}
In multi-task learning, the goal is to learn models for multiple
related tasks by taking advantage of pair-wise relationships between the tasks
(see, \eg, \cite{Caruana1997,ZY:17}).
Multi-task learning can be interpreted as a stratified model where the
stratification feature is the task to be performed by that model.
Laplacian regularization has also been investigated in the context of
multi-task learning \cite{sheldon2008graphical}.

\paragraph{Laplacian systems.}
Laplacian matrices are very well studied in the field of spectral
graph theory \cite{spielman2007spectral,teng2010laplacian}.
A set of linear equations with Laplacian coefficient matrix is 
called a \emph{Laplacian system}.
Many problems give rise to Laplacian systems, including in
electrical engineering (resistor networks), physics (vibrations and heat),
computer science, and spectral graph theory (see, \eg, \cite{spielman2010algorithms} for
a survey).
While solving a general linear system requires order $K^3$ flops,
where $K$ is the number of vertices in the graph, Laplacian system solvers
have been developed that are (nearly) linear in the number of edges contained in
the associated graph \cite{vishnoi2013lx,CKMPP:14,STvDD:17}.
In many cases, however, the graph can be large and dense enough that solving a Laplacian
system is still a highly nontrivial computational issue; in these cases, graph
sparsification can be used to yield results of moderate accuracy in much less time
\cite{sadhanala2016laplacian}.
In practice, the conjugate gradient (CG) method \cite{hestenes1952methods}, with a
suitable pre-conditioner, can be extremely efficient for solving the (positive definite)
Tikhonov-regularized Laplacian systems that we encounter in this paper.

\paragraph{Laplacian regularized minimization problems.}
Stratified model fitting with Laplacian regularization is
simply a convex optimization problem with Laplacian regularization.
There are many general-purpose methods that solve this problem, indeed too many to name;
in this paper we derive a problem-specific algorithm using ADMM.
Another possibility would be majorization-minimization (MM),
an algorithmic framework where one iteratively 
minimizes a majorizer of the original function at the current iterate \cite{SBP:17}.
One possible majorizer for the Laplacian regularization term is a diagonal quadratic,
allowing block separable problems to be solved in parallel; this was done in
\cite{tuck2018distributed}. The MM and ADMM algorithms have been shown to be
closely connected \cite{LFYL:18}, since  the proximal operator of a function
minimizes a quadratic majorizer. However, convergence with MM depends heavily
on the choice of majorizer and the problem structure \cite{mm_tutorial:04}.
In contrast, the ADMM-based algorithm we describe in this paper 
typically converges (at least to reasonable practical tolerance) within
100--200 iterations.

\section{Stratified models}\label{s-stratified_models}

In this section we define stratified models and our fitting method; we describe our 
distributed method for carrying out the fitting in \S\ref{s-distr-method}.

We consider data fitting problems with items or records of the 
form $(z,x,y) \in \mathcal Z \times \mathcal X \times \mathcal Y$.
Here $z\in \mathcal Z$ is the feature over which we stratify, 
$x\in \mathcal X$ is the other features, and $y\in \mathcal Y$ is the outcome or
dependent variable.
For our purposes, it does not matter what $\mathcal X$ or $\mathcal Y$ are;
they can consist of numerical, categorical, or other data types.
The stratified feature values $\mathcal Z$, however, 
must consist of only $K$ possible values, which we denote as $\mathcal Z =\{1, \ldots, K\}$.
What we call $z$ can include several original features, for example,
sex and age.  If all combinations of these original features are possible, then
$K$ is the product of the numbers of values these features can take on.
In some formulations, described below, we do not have $x$; in this case, the records
have the simplified form $(z,y)$.

\paragraph{Base model.}
We build a stratified model on top of a base model, which models pairs $(x,y)$
(or, when $x$ is absent, just $y$).
The base model is parametrized by a parameter vector $\theta \in \Theta \subseteq \reals^n$.
In a stratified model, we use a different value of the
parameter $\theta$ for each value of $z$.  We denote these
parameters as $\theta_1, \ldots, \theta_K$, where $\theta_k$ is the parameter
value used when $z=k$.
We let $\theta=(\theta_1,\ldots, \theta_K) \in \reals^{Kn}$
denote the collection of parameter values,
\ie, the parameter value for the stratified model.

\paragraph{Local loss and regularization.}
Let $(z_i,x_i,y_i)$, $i=1\ldots, N$,
denote a set of $N$ training data points or examples.
We will use regularized empirical loss minimization to choose the
parameters $\theta_1, \ldots, \theta_K$.
Let $l: \Theta \times \mathcal X \times \mathcal Y \to \reals$ be a loss function.
We define the $k$th (local) \emph{empirical loss} as
\BEQ\label{e-loc-obj}
\ell_k(\theta) = \sum_{i:z_i=k} l(\theta,x_i,y_i)
\EEQ
(without $x_i$ when $x$ is absent).

We may additionally add a regularizer to the local loss function.
The idea of regularization can be traced back to as early as the 1940s,
where it was introduced as a method to stabilize ill-posed 
problems \cite{tikhonov1943inverse,tikhonov1963regularization}.
Let $r: \Theta \to \reals \cup \{\infty\}$
be a regularization function or regularizer.
Choosing $\theta_k$ to minimize $\ell_k(\theta_k)+r(\theta_k)$ gives the
regularized empirical risk minimization model parameters, based only on the
data records that take the particular value of the stratification feature $z=k$.
This corresponds to the traditional
stratified model, with no connection between the parameter values for different
values of $z$.
(Infinite values of the regularizer encode constraints on allowable
model parameters.)

\paragraph{Laplacian regularization.}
Let $W \in \reals^{K\times K}$ be a symmetric matrix with nonnegative entries.
The associated \emph{Laplacian regularization} is the function
$\mathcal L: \reals^{Kn} \to \reals$ given by
\BEQ\label{e-lapl}
\mathcal L(\theta)
= \mathcal L(\theta_1, \ldots, \theta_K)
= \frac{1}{2}\sum_{i,j=1}^K W_{ij} \|\theta_i - \theta_j\|_2^2.
\EEQ
(To distinguish the Laplacian regularization $\mathcal L$ from $r$ in~(\ref{e-loc-obj}),
we refer to $r$ as the \emph{local regularization}.)
We can associate the Laplacian regularization with a graph with
$K$ vertices, which has an edge $(i,j)$ for
each positive $W_{ij}$, with weight $W_{ij}$.
We refer to this graph as the regularization graph.
We can express the Laplacian regularization as the positive semidefinite quadratic form
\[
\mathcal L(\theta) = (1/2) \theta^T (I \otimes L) \theta,
\]
where $\otimes$ denotes the Kronecker product,
and $L \in \reals^{K \times K}$ is the (weighted) Laplacian matrix associated with the
weighted graph, given by
\[
L_{ij}  = \left\{ \begin{array}{ll}
-W_{ij} & i \neq j \\
\sum_{k=1}^K W_{ik} & i=j
\end{array} \right.
\]
for $i,j=1,\ldots, K$.

The Laplacian regularization $\mathcal L(\theta)$ evidently measures the aggregate
deviation of the parameter vectors from their graph neighbors, weighted by the edge
weights.
Roughly speaking, it is a metric of how rough or non-smooth the mapping from $z$ to
$\theta_z$ is, measured by the edge weights.

\paragraph{Fitting the stratified model.}
To choose the parameters $\theta_1, \ldots, \theta_K$, we minimize
\BEQ
F(\theta_1, \ldots, \theta_k)= \sum_{k=1}^K \left( \ell_k(\theta_k) +
r (\theta_k) \right) +
\mathcal L(\theta_1, \ldots,\theta_K).
\label{e-strat-obj}
\EEQ
The first term is the sum of the local objective functions, used in fitting
a traditional stratified model;
the second measures the non-smoothness of the model parameters as measured by the
Laplacian.
It encourages parameter values for neighboring values of $z$ to be close to
each other.

\paragraph{Convexity assumption.}
We will assume that the local loss function $l(\theta,x,y)$ and the local regularizer
$r(\theta)$ are convex functions of $\theta$, which implies that
that the local losses $\ell_k$ are convex functions of $\theta$.
The Laplacian is also convex, so the overall objective $F$ is convex,
and minimizing it (\ie, fitting a stratified model) is a
convex optimization problem.
In \S\ref{s-distr-method} we will describe an effective distributed method
for solving this problem.
(Much of our development carries over to the case when $l$ or $r$ is not convex;
in this case, of course, we can only expect to find a local minimizer of $F$;
see \S\ref{s-extensions}.)

\paragraph{The two extremes.}
Assuming the regularization graph is connected,
taking the positive weights $W_{ij}$ to $\infty$ forces all the parameters to be equal,
\ie, $\theta_1 = \cdots = \theta_K$.  We refer to this as the \emph{common model}
(\ie, one which does not depend on $z$).
At the other extreme we can take all $W_{ij}$ to $0$,  which results in 
a traditional stratified model, with separate models independently fit for each 
value of $z$.

\paragraph{Hyper-parameters.}
The local regularizer and the weight matrix $W$ typically contain
some positive hyper-parameters, for example that scale 
the local regularization or one or more edges weights in the graph.
As usual, these are varied over a range of values, and for each value a 
stratified model is found, and tested on a separate validation data set using an
appropriate true objective.
We choose values of the 
hyper-parameters that give good validation set performance; finally,
we test this model on a new test data set.

\paragraph{Modeling with no data at nodes.}
Suppose that at a particular node on the graph (\ie, value of 
stratification feature), there is no data to fit a model.
Then, the model for that node will simply be a weighted average of the models of its
neighboring nodes.

\paragraph{Feature engineering interpretation.}
We consider an interpretation of stratified model fitting versus typical
model fitting in a linear model setting.
Instead of being a scalar in $\{1, \ldots, K\}$, let $z$ be a one-hot vector in $\reals^K$
where the data record is in the $k$th category if $z = e_k$, the $k$th basis vector in $\reals^k$.
In a typical model fitting formulation with inputs $x,z$ and outcomes $y$
an estimate of $y$ is given by $\hat y = (x, z)^T \theta$ with $\theta \in \reals^{n+K}$; 
that is, the model is \emph{linear} in $x$ and $z$.
With stratified model fitting, our estimate is instead
$\hat y = x^T \theta z$ with $\theta \in \reals^{n \times K}$; 
that is, the model is \emph{bilinear} in $x$ and $z$.

\subsection{Data models}
\label{sec:data-models}

So far we have been vague about what exactly a data model is.  
In this section we list just a few of the many possibilities.

\subsubsection{Point estimates}

The first type of data model that we consider is one where we wish to predict 
a single likely value, or point estimate, of $y$ for a given $x$ and $z$.
In other words, we want to construct a function $f_z: \mathcal X \to \mathcal Y$ such 
that $f_z(x) \approx y$.

\paragraph{Regression.}
In a regression data model, $\mathcal X = \reals^n$ and $\mathcal Y = \reals$,
with one component of $x$ always
having the value one.  The loss has the form $l(\theta, x,y) = p(x^T \theta - y)$,
where $p: \reals \to \reals$ is a penalty function.
Some common (convex) choices include the square penalty
\cite{legendre1805nouvelles,gauss1809theoria},
absolute value penalty \cite{boscovich1757litteraria},
Huber penalty \cite{huber1964robust},
and tilted absolute value (for quantile regression).
Common generic regularizers include zero, sum of squares,
and the $\ell_1$ norm \cite{tibshirani1996lasso}; these
typically do not include the coefficient in $\theta$
associated with the constant feature value one,
and are scaled by a positive hyper-parameter.
Various combinations of these choices lead to
ordinary least squares regression, ridge regression,
the lasso, Huber regression, and quantile regression.
Multiple regularizers can also be employed, each with a different
hyper-parameter, as in the elastic net.
Constraints on model parameters (for example, nonnegativity) 
can be imposed by infinite values of the regularizer.

The stratified regression model gives the predictor of $y$, given $z$ and $x$, given by
\[
f_z(x) = x^T \theta_z.
\]
The stratified model uses a different set of regression coefficients for each value of $z$.
For example, we can have a stratified lasso model, or a stratified 
nonnegative least squares model.

An interesting extension of the regression model, sometimes called multivariate 
or multi-task regression, takes $\mathcal Y = \reals^m$ with $m>1$.
In this case we take $\Theta = \reals^{n \times m}$, so $\theta_z$ is a matrix,
with $f_z(x) = x^T \theta_z$ as our predictor for $y$.
An interesting regularizer for multi-task regression is the nuclear norm,
which is the sum of singular values, scaled by a hyper-parameter
\cite{vandenberghe1996semidefinite}.  
This encourages the matrices $\theta_z$ to have low rank.

\paragraph{Boolean classification.}
In a boolean classification data model,
$\mathcal X = \reals^n$ and $\mathcal Y= \{-1,1\}$.
The loss has the form
$l(\theta,x,y)  = p(yx^T\theta)$,
where $p:\reals \to \reals$ is a penalty function. 
Common (convex) choices of $p$ are $p(u)=(u-1)^2$ (square loss);
$p(u) = (u-1)_+$ (hinge loss), and $p(u)=\log(1+\exp -u)$ (logistic loss).
The predictor is 
\[
f_z(x) = \sign(x^T\theta_z).
\]
The same regularizers mentioned in regression can be used.
Various choices of these loss functions and regularizers lead to least squares
classification, support vector machines (SVM) \cite{cortes1995svm}, 
logistic regression \cite{cox1958logistic,hosmer2005logisticbook}, and so on
\cite{hastie2009elements}.
For example, a stratified SVM model develops an SVM model for each
value of $z$; the Laplacian regularization encourages the parameters 
associated with neighboring values of $z$ to be close.

\paragraph{Multi-class classification.}
Here $\mathcal X = \reals^n$, and $\mathcal Y$ is finite, say, $\{1, \ldots, M\}$.
We usually take $\Theta = \reals^{n \times M}$, with associated predictor
$f_z(x) = \argmax_i \left( x^T \theta_z \right)_i$.
Common (convex) loss functions are the multi-class logistic (multinomial) 
loss \cite{engel1988multinomiallogistic},
\[
l(\theta,x,y) = \log\left(\sum_{j=1}^M \exp (x^T \theta)_j \right) - (x^T \theta)_y,
\quad y=1, \ldots, M,
\]
and the multi-class SVM loss \cite{Yang2007multiclassSVM,manning2008multiclassSVM},
\[
l(\theta,x,y) = \sum_{i:i\neq y}((x^T \theta)_i - (x^T \theta)_y + 1)_+, \quad y=1, \ldots, M.
\]
The same regularizers mentioned in multivariate regression can be used.

\paragraph{Point estimates without $x$.}

As mentioned above, $x$ can be absent.
In this case, we predict $y$ given only $z$.
We can also interpret this as a point estimate with
an $x$ that is always constant, \eg, $x=1$.
In a regression problem without $x$,
the parameter $\theta_z$ is a scalar, and
corresponds simply to a prediction of the value $y$.
The loss has the form $l(\theta,y)=p(\theta-y)$,
where $p$ is any of the penalty functions 
mentioned in the regression section.
If, for example, $p$ were the square (absolute value) penalty, then $\theta_k$
would correspond to 
the average (median) of the $y_i$ that take on the particular stratification
feature $k$, or $z_i=k$.
In a boolean classification problem, $\theta_z \in \reals$, and the predictor is
$\sign(\theta_z)$, 
the most likely class.
The loss has the form $l(\theta,y) = p(y\theta)$, where $p$
is any of the penalty functions 
mentioned in the boolean classification section.
A similar predictor can be derived for the case of multi-class classification.

\subsubsection{Conditional distribution estimates}\label{ssec:conditional_dist_est}

A more sophisticated data model predicts
the conditional probability distribution of $y$ given $x$ (rather than a specific
value of $y$ given $x$, in a point estimate).
We parametrize this conditional probability distribution by a vector $\theta$,
which we denote as $\Prob(y \mid x, \theta)$.
The data model could be any generalized linear model
\cite{nelder1972generalized}; here we describe a few common ones.

\paragraph{Logistic regression.}\label{ss:logreg}
Here $\mathcal X = \reals^n$ and $\mathcal Y = \{-1,1\}$, with one
component of $x$ always
having the value one.
The conditional probability has the form
\[
\Prob(y = 1 \mid x, \theta) = \frac{1}{1+e^{-x^T\theta}}.
\]
(So $\Prob(y = -1 \mid x, \theta) = 1-\Prob(y = 1 \mid x, \theta)$.)
The loss, often called the logistic loss,
has the form $l(\theta, x, y) = \log(1+\exp(-y\theta^T x))$,
and corresponds to the negative log-likelihood of the data record $(x,y)$,
when the parameter vector is $\theta$.
The same regularizers mentioned in regression can be used.
Validating a logistic regression data model using the average logistic
loss on a test set is the same (up to a constant factor)
as the average log probability of
the observed values in the test data, under the predicted
conditional distributions.

\paragraph{Multinomial logistic regression.}
Here $\mathcal X = \reals^n$ and $\mathcal Y$ is finite, say $\{1,\ldots,M\}$
with one component of $x$ equal to one.
The parameter vector is a matrix $\theta \in \reals^{n \times M}$.
The conditional probability has the form
\[
\Prob(y = i \mid x, \theta) = \frac{\exp (x^T \theta)_i}{\sum_{j=1}^M \exp (x^T \theta)_j},
\quad i=1, \ldots, M.
\]
The loss has the form
\[
l(\theta, x, y=i) = \log\left(\sum_{j=1}^M \exp (x^T \theta)_j \right) - (x^T \theta)_i,
\quad i=1, \ldots, M,
\]
and corresponds to the negative log-likelihood of the data record $(x,y)$.

\paragraph{Exponential regression.}
Here $\mathcal X = \reals^n$ and
$\mathcal Y = \reals_+$ (\ie, the nonnegative reals).
The conditional probability distribution in exponential regression has the form
\[
\Prob(y \mid x, \theta) = \exp(x^T\theta) e^{-\exp(x^T\theta) y},
\]
where $\theta\in\reals^n$ is the parameter vector.
This is an exponential distribution over $y$, with
rate parameter given by $\omega=\exp(x^T\theta)$.
The loss has the form
\[
l(\theta, x, y) =  - x^T\theta + \exp(x^T\theta) y,
\]
which corresponds to the negative log-likelihood of the data record
$(x,y)$ under the exponential distribution with parameter $\omega = \exp(x^T\theta)$.
Exponential regression is useful when the outcome $y$ is positive.
A similar conditional probability distribution can be derived
for the other distributions, \eg, the Poisson distribution. 

\subsubsection{Distribution estimates}
\label{sec:distribution_estimates}

A distribution estimate is an estimate of the probability distribution of $y$
when $x$ is absent, denoted $p (y\mid\theta)$.
This data model can also be interpreted as a conditional distribution estimate
when $x=1$.
The stratified distribution model consists of a separate distribution for $y$,
for each value of the stratification parameter $z$.
The Laplacian regularization in this case encourages closeness between
distributions that have similar stratification parameters, measured by the
distance between their model parameters.

\paragraph{Gaussian distribution.}
Here $\mathcal Y = \reals^m$, and we fit a density to the observed values 
of $y$, given $z$.
For example $\theta$ can parametrize a Gaussian on $\reals^m$, $\mathcal N(\mu,\Sigma)$.
The standard parametrization uses the parameters $\theta=(\Sigma^{-1},\Sigma^{-1}\mu)$,
with $\Theta = \symm_{++}^m \times \reals^m$
($\symm_{++}^n$ denotes the set of $n \times n$ positive definite matrices.)
The probability density function of $y$ is
\[
p(y\mid\theta) = (2\pi)^{-m/2}\det(\Sigma)^{-1/2}\exp(-\frac{1}{2}(y-\mu)^T
\Sigma^{-1}(y-\mu))
\]
The loss function (in the standard parametrization) is
\[
l(\theta,y) = -\log\det S + y^TSy -2y^T\nu+\nu^TS^{-1}\nu
\]
where $\theta=(S,\nu)=(\Sigma^{-1},\Sigma^{-1}\mu)$.
This loss function is jointly convex in $S$ and $\nu$;
the first three terms are evidently convex and the fourth is
the matrix fractional function (see \cite[p76]{boyd2004convex}).
Some common (convex) choices for the local regularization function
include the trace of $\Sigma^{-1}$ (encourages the harmonic mean of
the eigenvalues of $\Sigma$, and hence volume, to be large),
sum of squares of $\Sigma^{-1}$ (shrinks the overall conditional dependence
between the variables), $\ell_1$-norm of $\Sigma^{-1}$ (encourages
conditional independence between the variables)
\cite{friedman2008sparse},
or a prior on $\Sigma^{-1}$ or $\Sigma^{-1} \mu$.

When $m=1$, this model corresponds to the standard normal distribution,
with mean $\mu$ and variance $\sigma^2$.
A stratified Gaussian distribution model has a separate
mean and covariance of $y$ for each value of the stratification parameter $z$.
This model can be validated by evaluating the average log density of
the observed values $y_i$, under the predicted distributions, over a test set of data.

\paragraph{Bernoulli distribution.}\label{ss:bernoulli}
Here $\mathcal Y = \reals^n_+$.
The probability density function of a Bernoulli distribution has the form
\[
p(y \mid \theta) = \theta^{\ones^T y} (1-\theta)^{n - \ones^T y},
\]
where $\theta \in \Theta = [0,1]$. 
The loss function has the form
\[
l(\theta, y) = -(\ones^T y) \log(\theta) - (n - \ones^T y) \log(1-\theta),
\]
which corresponds to the negative log-likelihood of the Bernoulli distribution.

\paragraph{Poisson distribution.}
Here $\mathcal Y = \mathbb{Z}_+$ (\ie, the nonnegative integers).
The probability density function of a Poisson distribution has the form
\[
p(y=k\mid\theta) = \frac{\theta^k e^{-\theta}}{k!}, \quad k=0,1,\ldots,
\]
where $\theta \in \Theta = \reals_+$.
If $\theta=0$, then $p(y=0\mid \theta) = 1$.
The loss function has the form
\[
l(\theta,y=k) = -k\log\theta + \theta, \quad k=0,1,\ldots,
\]
which corresponds to the negative log-likelihood of the Poisson distribution.
A similar data model can be derived for the exponential distribution.

\paragraph{Non-parametric discrete distribution.}

Here $\mathcal Y$ is finite, say $\{1,\ldots,M\}$ where $M>1$.
We fit a non-parametric discrete distribution to $y$, which has the form
\[
p(y = k \mid \theta) = \theta_k, \quad k=1,\ldots,M,
\]
where $\theta\in\Theta = \{p\in\reals^M \mid \ones^Tp=1, ~p \succeq 0\}$, \ie,
the probability simplex.
The loss function has the form
\[
l(\theta,y=k) = -\log(\theta_k), \quad k=1,\ldots,M.
\]
A common (convex) regularization function is the negative entropy, given by
$\sum_{i=1}^M \theta_i \log(\theta_i)$.

\paragraph{Exponential families.}

A probability distribution that can be expressed as
\[
p(y \mid \theta) = e^{M(\theta,y)}
\]
where $M:\reals^n \times \mathcal Y \to \reals$ is a concave function of $\theta$, 
is an exponential family \cite{koopman1936exponentialfamily,pitman1936exponentialfamily}.
Some important special cases include the Bernoulli, multinomial, Gaussian, and 
Poisson distributions.
A probability distribution that has this form naturally leads to the following
(convex) loss function
\[
l(\theta,y) = -\log p(y \mid \theta) = -M(\theta,y).
\]

\subsection{Regularization graphs}

In this section we outline some common regularization graphs, which
are undirected weighted graphs where each vertex corresponds to a possible
value of $z$.
We also detail some possible uses of each type of regularization graph in
stratified model fitting.

\paragraph{Complete graph.}
A complete graph is a (fully connected) graph that contains every possible edge.
In this case, all of the models for each stratification feature are
encouraged to be close to each other.

\paragraph{Star graph.}
A star graph has one vertex with edges to every other vertex.
The vertex that is connected to every other vertex is sometimes called the
\emph{internal} vertex.
In a stratified model with a star regularization graph,
the parameters of all of the non-internal vertices are encouraged to be similar
to the parameter of the internal vertex.
We refer to the internal vertex as the \emph{common model}.
A common use of a star graph is when the stratification feature relates many
vertices only to the internal vertex.

It is possible to have no data associated with the common model or internal vertex.
In this case the common model parameter is a weighted average of other 
model parameters; it serves to pull the other parameter values together.

It is a simple exercise to show that a stratified model with the complete graph
can also be found using a star graph, 
with a central or internal vertex that is connected to all others, but has no data.
(That is, it corresponds to a new fictitious value of the 
stratification feature, that does not occur in the real data.)

\paragraph{Path graph.}
A path graph, or linear/chain graph, is a graph whose vertices
can be listed in order, with edges between adjacent vertices in that order.
The first and last vertices only have one edge, whereas the other vertices
have two edges.
Path graphs are a natural choice for when the stratification feature corresponds
to time, or location in one dimension.
When the stratification feature corresponds to time,
a stratified model with a path regularization graph correspond to a
time-varying model, where the model varies smoothly with time.

\paragraph{Cycle graph.}
A cycle graph or circular graph is a graph where the vertices
are connected in a closed chain.
Every vertex in a cycle graph has two edges.
A common use of a cycle graph is when the stratification feature
corresponds to a periodic variables, \eg, the day of the week;
in this case, we fit a separate model for each day of the week, and the model parameter 
for Sunday is close to both the model parameter for Monday 
and the model parameter for Saturday.
Other examples include days of the year, the season, and non-time variables such
as angle.

\paragraph{Tree graph.}
A tree graph is a graph where any two vertices are connected by exactly one
path (which may consist of multiple edges).
Tree graphs can be useful for stratification features that naturally have
hierarchical structure, \eg, the location in a corporate hierarchy, 
or in finance, hierarchical classification of individual stocks into
sectors, industries, and sub-industries.
As in the star graph, which is a special case of a tree graph, internal
vertices need not have any data associated with them, \ie, the data is only
at the leaves of the tree.

\paragraph{Grid graph.}

A grid graph is a graph where the vertices correspond to points
with integer coordinates, and two vertices are connected if their
maximum distance in each coordinate is less than or equal to one.
For example, a path graph is a one-dimensional grid graph.

Grid graphs are useful for when the stratification
feature corresponds to locations.
We discretize the locations into bins, and connect adjacent bins in each
dimension to create a grid graph.
A space-stratified model with grid graph regularization is encouraged
to have model parameters that vary smoothly through physical space.

\paragraph{Entity graph.}
There can be a vertex for every value of some entity, \eg, a person.
The entity graph has an edge between two entities if they are perceived to be
similar.
For example, two people could be connected in a (friendship) entity graph if they
were friends on a social networking site.
In this case, we would fit a separate model for each person, and encourage
friends to have similar models.
We can have multiple relations among the entities, each one associated with its
own graph; these would typically be weighted by separate hyper-parameters.

\paragraph{Products of graphs.}

As mentioned earlier, $z$ can include several original features, and $K$, 
the number of possible values of $z$, is the product of the number of values that 
these features take on.
Each original stratification feature can have its own regularization graph,
with the resulting 
regularization graph for $z$ given by the (weighted) Cartesian product of the 
graphs of each of the features.
For example, the product of two path graphs is a grid graph, with the horizontal
and vertical edges having different weights, associated with the two
original stratification features.

\section{Distributed method for stratified model fitting}
\label{s-distr-method}

In this section we describe a distributed algorithm
for solving the fitting problem \eqref{e-strat-obj}.
The method alternates between computing (in parallel) the proximal operator of the
local loss and local regularizer for each model, and computing the proximal operator 
of the Laplacian regularization term, 
which requires the solution of a number of regularized Laplacian linear systems
(which can also be done in parallel).

To derive the algorithm, we first express \eqref{e-strat-obj} in the equivalent form 
\BEQ
\label{e-variant}
\begin{aligned}
& \text{minimize}
& & \sum_{k=1}^K (\ell_k(\theta_k) + r({\tilde{\theta}}_k))
+ \mathcal L({\hat{\theta}}_1, \ldots, {\hat{\theta}}_K) \\
& \text{subject to}
&& \theta = \hat{\theta}, \quad \tilde{\theta} = \hat{\theta},
\end{aligned}
\EEQ
where we have introduced two additional
optimization variables $\tilde\theta\in \reals^{Kn}$
and $\hat\theta\in\reals^{Kn}$.
The augmented Lagrangian $L_{\lambda}$ of \eqref{e-variant} has the form
\BEAS
L_{\lambda}(\theta, \tilde{\theta}, \hat{\theta}, u, \tilde u) 
&=& \sum_{k=1}^K (\ell_k(\theta_k) + r(\tilde{\theta_k})) 
+ \mathcal L({\hat{\theta}}_1, \ldots, {\hat{\theta}}_K)\\
&& \mbox{}  + (1/2\lambda)\|\theta - \hat{\theta} + u\|_2^2
+ (1/2\lambda)\|\tilde{\theta} - \hat{\theta} + \tilde u\|_2^2,
\EEAS
where $u\in\reals^{Kn}$ and $\tilde u\in\reals^{Kn}$
are the (scaled) dual variables associated 
with the two constraints in \eqref{e-variant},
respectively, and $\lambda>0$ is the penalty parameter.
The ADMM algorithm (in scaled dual form) for
the splitting $(\theta,\tilde\theta)$ and
$\hat\theta$ consists of the iterations
\BEAS
\theta^{i+1}, \tilde\theta^{i+1}
&\coloneqq&
\argmin_{\theta,\tilde\theta}
L_\lambda(\theta, \tilde\theta, \hat\theta^i, u^i, \tilde u^i) \\
\hat\theta^{i+1} &\coloneqq&
\argmin_{\hat\theta} L_\lambda(\theta^i, \tilde\theta^i, \hat\theta, u^i, \tilde u^i)\\
u^{i+1} &\coloneqq& u^i + \theta^{i+1} - {\hat{\theta}}^{i+1} \\
\tilde u^{i+1} &\coloneqq& {\tilde{u}}^i + \tilde{\theta}^{i+1} - {\hat{\theta}}^{i+1}.
\EEAS
Since the problem is convex, the iterates $\theta^i$, $\tilde\theta^i$, and $\hat\theta^i$
are guaranteed to converge to each other and to a primal optimal point of \eqref{e-strat-obj}
\cite{boyd2011distributed}.

This algorithm can be greatly simplified (and parallelized) with a few simple observations.
Our first observation is that the first step in ADMM can be expressed as
\[
\theta^{i+1}_k = \prox_{\lambda l_k}(\hat\theta_k^i-u_k^i), \quad
\tilde\theta^{i+1}_k = \prox_{\lambda r}(\hat\theta_k^i-\tilde u_k^i), \quad
k=1,\ldots,K,
\]
where $\prox_{g}: \reals^n \to \reals^n$ is the
proximal operator of the function $g$ \cite{PB:14} 
(see the Appendix for the definition and some examples of proximal operators).
This means that we can compute $\theta^{i+1}$
and $\tilde\theta^{i+1}$ at the same time, since they do not 
depend on each other.
Also, we can compute $\theta_1^{i+1},\ldots,\theta_K^{i+1}$ 
(and $\tilde\theta_1^{i+1},\ldots,\tilde\theta_K^{i+1}$) in parallel.

Our second observation is that the second step in ADMM can be expressed as the
solution to the $n$ regularized Laplacian systems
\BEQ
\label{e-laplacian_system}
\bigg(L+ (2/\lambda) I\bigg)
\begin{bmatrix}
(\hat\theta^{i+1}_1)_j \\
(\hat\theta^{i+1}_2)_j \\
\vdots \\
(\hat\theta^{i+1}_K)_j
\end{bmatrix}
= (1/\lambda)
\begin{bmatrix}
(\theta^{i+1}_1 + u^{i}_1 + {\tilde\theta}^{i+1}_1 + {\tilde u}^{i}_1)_j \\
(\theta^{i+1}_2 + u^{i}_2 + {\tilde\theta}^{i+1}_2 + {\tilde u}^{i}_2)_j \\
\vdots \\
(\theta^{i+1}_K + u^{i}_K + {\tilde\theta}^{i+1}_K + {\tilde u}^{i}_K)_j
\end{bmatrix},
\quad
j=1,\ldots,n.
\EEQ
These systems can be solved in parallel.
Many efficient methods for solving these systems have been developed;
we find that the conjugate gradient (CG) method \cite{HS:52}, 
with a diagonal pre-conditioner, can efficiently and reliably solve
these systems.
(We can also warm-start CG with $\hat\theta^i$.)
Combining these observations leads to Algorithm \ref{a-dist_strat}.

\begin{algdesc}
\label{a-dist_strat}
\emph{Distributed method for fitting stratified models with Laplacian regularization.}
\begin{tabbing}
    {\bf given} Loss functions $\ell_1,\ldots,\ell_K$, 
    local regularization function $r$,\\
    graph Laplacian matrix $L$, and penalty parameter $\lambda>0$.\\
    \emph{Initialize}.
    $\theta^0={\tilde\theta}^0={\hat{\theta}}^0=u^0={\tilde u}^0=0$.\\
    {\bf repeat} \\
    \qquad \=\ \bf{in parallel}\\
    \qquad \qquad \=\ 1.\
    \emph{Evaluate proximal operator of $l_k$.}
    $\theta^{i+1}_k = \prox_{\lambda \ell_k}({\hat{\theta}}_k^i - u_k^i), 
    \quad k = 1,\ldots,K$\\
    \qquad \qquad \=\ 2.\
    \emph{Evaluate proximal operator of $r$.}
    ${\tilde\theta}^{i+1}_k 
    = \prox_{\lambda r}({\hat{\theta}}_k^i - {\tilde u}_k^i), 
    \quad k = 1,\ldots,K$\\
    \qquad \=\ 3.\
    Solve the regularized Laplacian systems \eqref{e-laplacian_system} in parallel.
    \\
    \qquad \=\ 4.\
    \emph{Update the dual variables.}
    ${{u}}^{i+1} \coloneqq {{u}}^i + {\theta}^{i+1} - {\hat{\theta}}^{i+1}; 
    \quad {\tilde{u}}^{i+1}
    \coloneqq {\tilde{u}}^i + \tilde{\theta}^{i+1} - {\hat{\theta}}^{i+1}$\\
    {\bf until convergence}
\end{tabbing}
\end{algdesc}

\paragraph{Evaluating the proximal operator of $l_k$.}
Evaluating the proximal operator of $l_k$ corresponds to solving
a fitting problem with sum-of-squares regularization.
In some cases, this has a closed-form expression, and in others, it
requires solving a small convex optimization problem.
When the loss is differentiable, we can use, for example, the L-BFGS 
algorithm to solve it \cite{liu1989limited}.
(We can also warm start with $\theta^i$.)
We can also re-use factorizations of matrices (\eg, Hessians) used in
previous evaluations of the proximal operator.

\paragraph{Evaluating the proximal operator of $r$.}
The proximal operator of $r$ often has a closed-form expression.
For example, if $r(\theta)=0$, then the proximal operator of $r$ corresponds
to the projection onto $\Theta$.
If $r$ is the sum of squares function and
$\Theta = \reals^n$, then the proximal operator of
$r$ corresponds to the (linear) shrinkage operator.
If $r$ is the $\ell_1$ norm and $\Theta = \reals^n$, then the proximal operator of
$r$ corresponds to soft-thresholding, which can be performed in
parallel.

\paragraph{Stopping criterion.}
The primal and dual residuals
\[
r^{i+1} = (\theta^{i+1}-\hat\theta^{i+1},\tilde\theta^{i+1}-\hat\theta^{i+1}),\quad
s^{i+1} = -(1/\lambda)(\hat\theta^{i+1}-\hat\theta^{i},\hat\theta^{i+1}-\hat\theta^{i}),
\]
converge to zero \cite{boyd2011distributed}.
This suggests the stopping criterion
\[
\|r^{i+1}\|_2 \leq \epsilon_\mathrm{pri}, \quad \|s^{i+1}\|_2 \leq \epsilon_\mathrm{dual},
\]
where $\epsilon_\mathrm{pri}$ and $\epsilon_\mathrm{dual}$ are given by
\[
\epsilon_\mathrm{pri} 
= \sqrt{2Kn}\epsilon_\mathrm{abs} + \epsilon_\mathrm{rel} \max\{\|r^{i+1}\|_2,\|s^{i+1}\|_2\},
\quad \epsilon_\mathrm{dual} 
= \sqrt{2Kn}\epsilon_\mathrm{abs} + (\epsilon_\mathrm{rel}/\lambda) \|(u^i,{\tilde u^i})\|_2,
\]
for some absolute tolerance $\epsilon_{\mathrm{abs}} > 0$
and relative tolerance $\epsilon_{\mathrm{rel}} > 0$.

\paragraph{Selecting the penalty parameter.}
Algorithm \ref{a-dist_strat} will converge regardless of the choice of the penalty parameter
$\lambda$.
However, the choice of the penalty parameter can affect the speed of convergence.
We adopt the simple adaptive scheme \cite{he2000alternating,wang2001decomposition}
\[
\lambda^{k+1} \coloneqq \begin{cases}
\lambda^k/\tau^\mathrm{incr} & \text{if} \; \|r^k\|_2 > \mu\|s^k\|_2 \\
\tau^\mathrm{decr} \lambda^k & \text{if} \; \|s^k\|_2 > \mu\|r^k\|_2 \\
\lambda^k & \text{otherwise},
\end{cases}
\]
where $\mu>1$, $\tau^\mathrm{incr} > 1$, and $\tau^\mathrm{decr} > 1$ are parameters.
(We recommend $\mu=5$, $\tau^\mathrm{incr}=2$, and $\tau^\mathrm{decr}=2$.)
We found that this simple scheme with $\lambda^0=1$
worked very well across all of our experiments.
When $\lambda^{i+1} \neq \lambda^{i}$, we must re-scale the dual variables,
since we are using with scaled dual variables.
The re-scaling is given by
\[
u^{i+1} = c u^{i+1}, \quad \tilde u^{i+1} = c \tilde u^{i+1},
\]
where $c=\lambda^{i+1}/\lambda^{i}$.

\paragraph{Regularization path via warm-start.}
Our algorithm supports \emph{warm starting} by choosing the initial 
point $\theta^0$ as an estimate of the solution,
for example, the solution of a closely related problem (\eg, a problem with
slightly varying hyper-parameters.)

\paragraph{Software implementation.}

We provide an (easily extensible) implementation of the ideas described in the paper,
available at \url{www.github.com/cvxgrp/strat_models}.
We use \verb|numpy| for dense matrix representation and operations \cite{van2011numpy},
\verb|scipy| for sparse matrix operations and statistics functions \cite{jones2001scipy},
\verb|networkx| for graph factory functions and
Laplacian matrix computation \cite{hagberg2008exploring},
\verb|torch| for L-BFGS and GPU computation \cite{paszke2017automatic},
and \verb|multiprocessing| for parallel processing.
We provide implementations for local loss and local regularizers for creating a custom 
stratified model.
A stratified model can be instantiated with the following code:
\begin{equation*}
\begin{array}{ll}
\verb|base_model = strat_models.BaseModel(loss,reg)|\\
\verb|model = strat_models.StratifiedModel(base_model, G)|.
\end{array}
\end{equation*}
Here, $\verb|loss|$ and $\verb|reg|$ are implementations for the local loss and 
local regularizer, respectively, and $\verb|G|$ is a \verb|networkx| graph with nodes 
corresponding to $\mathcal Z$ describing the Laplacian.
The stratified model class supports the method
\[
\verb|model.fit(data)|,
\]
where $\verb|data|$ is a Python dictionary with with the data records $(x,y,z)$
(or just $(y,z)$ for distribution estimates).
We also provide methods that automatically compute cross validation scores.

We note that our implementation is for the most part expository
and for experimentation with stratified models;
its speed could evidently be improved by using a compiled language (\eg, C or C++).
Practictioners looking to use these methods at a large scale
should develop specialized codes
for their particular application using the ideas that we have described.

\section{Examples}
\label{s-examples}

In this section we illustrate the effectiveness of stratified models
by combining base fitting methods and regularization graphs
to create stratified models.
We note that these examples are all highly simplified in order to illustrate
their usage; better models can be devised using the very same techniques
illustrated.
In each example, we fit three models:  
a stratified model without Laplacian regularization
(which we refer to as a \emph{separate model}), 
a \emph{common model} without stratification,
and a \emph{stratified model} with hand-picked edge weights.
(In practice the edge weights should be selected using
a validation procedure.)
We find that the stratified model
significantly outperforms the other two methods in each example.

The code is available online at
\url{https://github.com/cvxgrp/strat_models}.
All numerical experiments were performed on an
unloaded Intel i7-8700K CPU.

\subsection{Mesothelioma classification}
We consider the problem of predicting whether a patient has mesothelioma, a form of cancer,
given their sex, age, and other medical features 
that were gathered during a series of patient encounters and laboratory studies.

\paragraph{Dataset.}
We obtained data describing \num{324} patients from the Dicle University Faculty of Medicine
\cite{er2012mesothelioma,dua2019ucirepository}.
The dataset is comprised of males and females between 
(and including) the ages of \num{19} and \num{85}, with
\num{96} (29.6\%) of the patients diagnosed with mesothelioma.
The 32 medical features in this dataset
include: city, asbestos exposure, type of MM, duration of
asbestos exposure, diagnosis method, keep
side, cytology, duration of symptoms, dyspnoea, ache on chest,
weakness, habit of cigarette, performance status, white blood
cell count (WBC), hemoglobin (HGB), platelet count (PLT), sedimentation,
blood lactic dehydrogenise (LDH), alkaline phosphatise
(ALP), total protein, albumin, glucose, pleural lactic dehydrogenise,
pleural protein, pleural albumin, pleural glucose,
dead or not, pleural effusion, pleural thickness on tomography,
pleural level of acidity (pH), C-reactive protein (CRP), class of
diagnosis (whether or not the patient has mesothelioma).
We randomly split the data into a training set containing
90\% of the records, and a test set containing
the remaining 10\%.

\paragraph{Data records.}
The `diagnosis method' feature is perfectly correlated with the output, so we removed it (this was not done in many other studies on this dataset which led to near-perfect classifiers).
We performed rudimentary feature engineering on the raw medical features to derive a feature vector $x\in\reals^{46}$.
The outcomes $y \in \{0,1\}$ denote whether or not the patient
has mesothelioma, with $y=1$ meaning the patient has
mesothelioma.
The stratification feature $z$ is a tuple consisting of
the patient's sex and age;
for example, $z=(\text{Male}, 62)$ corresponds to a 62 year old male.
Here the number of stratification features 
$K = 2 \cdot 67 = 134$.

\paragraph{Data model.}
We model the conditional probability of contracting mesothelioma 
given the features using logistic regression, as described in
\S\ref{ss:logreg}.
We employ sum of squares regularization on the model parameters,
with weight $\gamma_\mathrm{local}$.

\paragraph{Regularization graph.}
We take the Cartesian product of two regularization graphs:
\begin{itemize}
    \item \emph{Sex}. 
    The regularization graph has one edge between male and female, 
    with edge weight $\gamma_\mathrm{sex}$.
    \item \emph{Age}.
    The regularization graph is a path graph between ages,
    with edge weight $\gamma_\mathrm{age}$.
\end{itemize}

\begin{table}
\caption{Mesothelioma results.}
  \vspace{.4em}
  \label{tab:mesothelioma}
  \centering
  \begin{tabular}{lll}
    \toprule
    Model & Test ANLL & Test error \\
    \midrule
    Separate & \num{1.31} & \num{0.45} \\
    \textbf{Stratified} &  \textbf{0.76} & \textbf{0.27} \\
    Common & \num{0.73} & \num{0.30} \\
    \bottomrule
  \end{tabular}
\end{table}

\begin{figure}
  \centering
    \includegraphics[width=\textwidth]{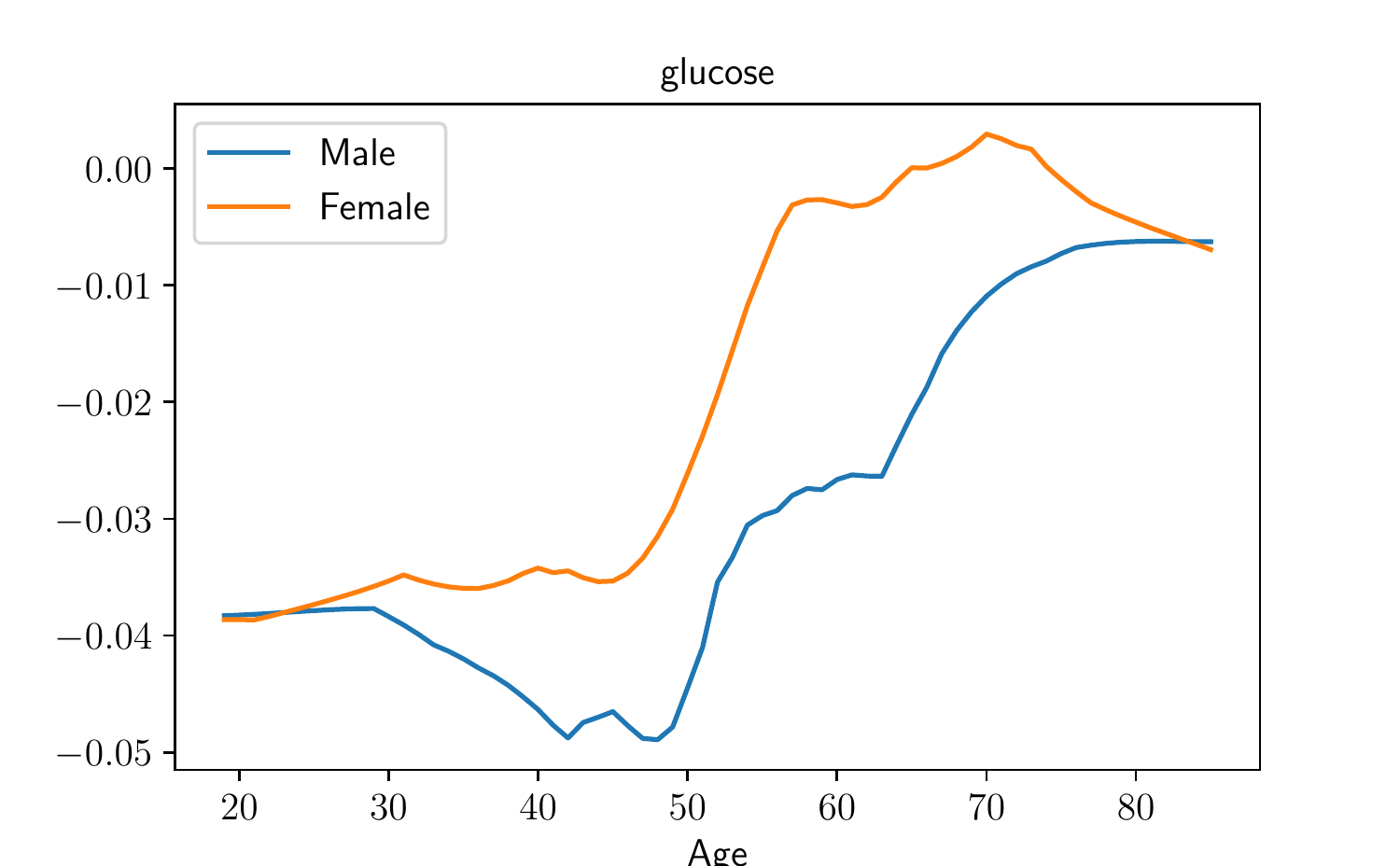}
      \caption{Glucose parameter versus age,
      for males and females.}
  \label{fig:mesothelioma_glucose}
\end{figure}

\paragraph{Results.}
We used $\gamma_\mathrm{local}=0.1$ for all experiments,
and $\gamma_\mathrm{sex}=10$ and $\gamma_\mathrm{age}=500$ for the stratified model.
We compare the average negative log likelihood (ANLL) 
and prediction error on the test set between
all three models in table \ref{tab:mesothelioma}.
The stratified model performs
slightly better at predicting the presence of
mesothelioma than the common model,
and much better than the separate model.
Figure \ref{fig:mesothelioma_glucose} displays the stratified model 
glucose parameter over age and sex.

\subsection{House price prediction}

We consider the problem of predicting the logarithm of a house's
sale price based on its geographical location
(given as latitude and longitude), and various features describing the house.

\paragraph{Data records.}
We gathered a dataset of sales data for homes in King County, WA
from May 2014 to May 2015.
Each record includes the latitude/longitude of the sale
and $n=10$ features: the number of bedrooms,
number of bathrooms, number of floors, waterfront (binary),
condition (1-5), grade (1-13), year the house was built (1900-2015),
square footage of living space,
square footage of the lot, and a constant feature (1).
We randomly split the dataset into 16197 training examples and
5399 test examples, and we standardized the features so that
they have zero mean and unit variance.

\paragraph{Data model.}
The data model here is ordinary regression with square loss.
We use the sum of squares local regularization function (excluding the
intercept) with regularization weight $\gamma_\mathrm{local}$.

\paragraph{Regularization graph.}
We binned latitude/longitude into $50 \times 50$ equally
sized bins.
The regularization graph here is a grid graph with edge weight
$\gamma_\mathrm{geo}$.

\begin{figure}
  \centering
    \includegraphics[width=\textwidth]{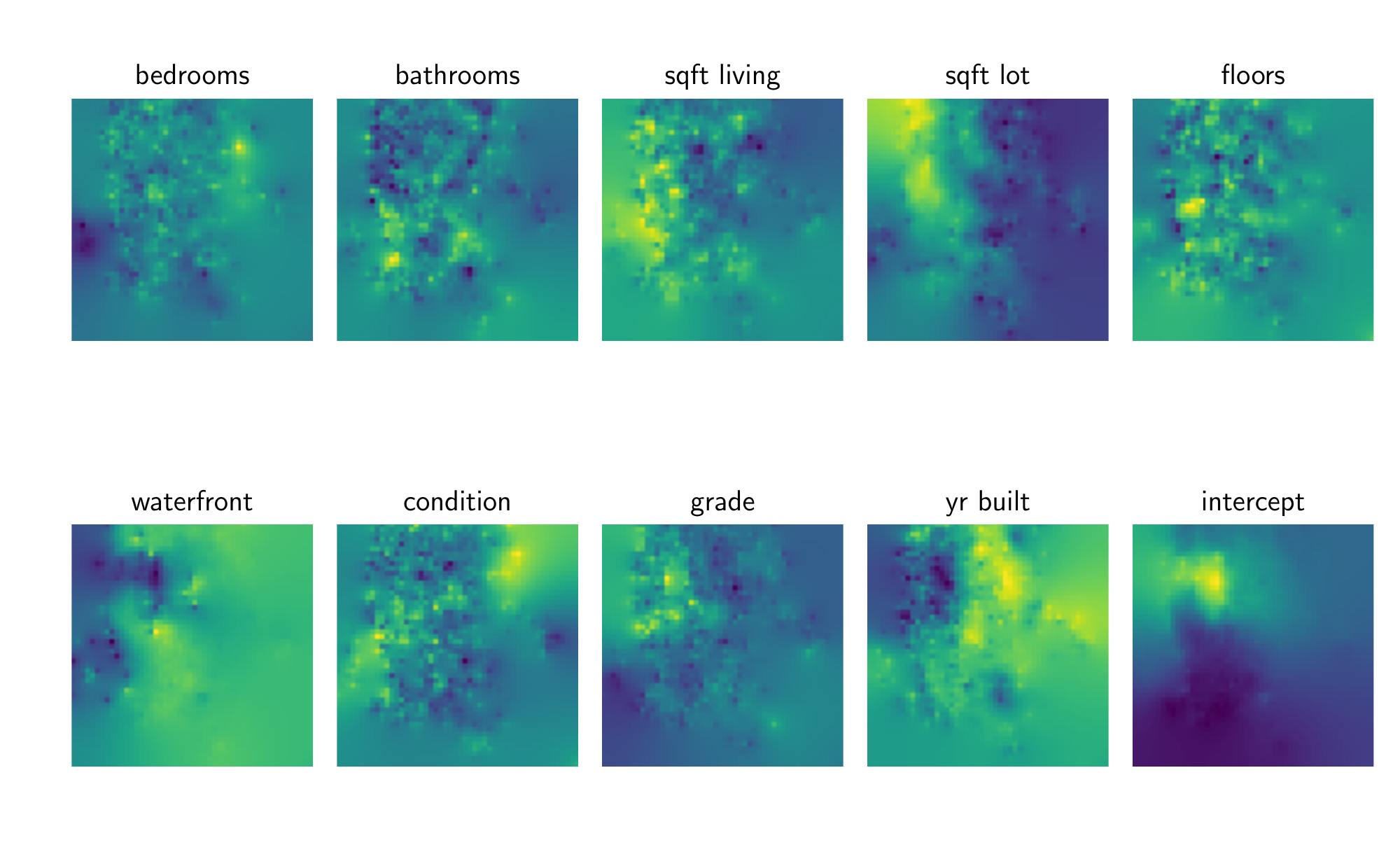}
      \caption{House price prediction coefficients.}
  \label{fig:house_price_coef}
\end{figure}

\begin{table}
  \caption{House price prediction results.}
  \vspace{.4em}
  \centering
  \begin{tabular}{lll}
    \toprule
    Model & No. parameters & Test RMSE \\
    \midrule
    Separate                & 22500             & 1.514 \\
    \textbf{Stratified}     & \textbf{22500}    & \textbf{0.181} \\
    Common                  & 10                & 0.316 \\
    Random forest           & 985888            & 0.184 \\
    \bottomrule
  \end{tabular}
  \label{tab:house}
\end{table}

\paragraph{Results.}
In all experiments, we used $\gamma_\mathrm{local}=1$.
(In practice this should be determined using cross-validation.)
We compared a stratified model ($\gamma_\mathrm{geo}=15$) to
a separate model,
a common model,
and a random forest regressor with 50 trees
(widely considered to be the best ``out of the box'' model).
We gave the latitude and longitude as raw features
to the random forest model.
In table \ref{tab:house}, we show the size of these models
(for the random forest this is the total number of nodes)
and their test RMSE.
The Laplacian regularized stratified model performs better
than the other models, and in particular, outperforms the random forest,
which has almost two orders of magnitude more parameters and is
much less interpretable.

In figure \ref{fig:house_price_coef}, we show the model parameters
for each feature across location.
In this visualization, the brightness of a particular location
can be interpreted as the influence that increasing
that feature would have on log sales price.
(A dark spot means that an increase in that feature in fact
decreases log sales price.)
For example, the waterfront feature seems to have a positive
effect on sales price only in particular coastal areas.

\subsection{Senate elections}
We model the probability that a
United States Senate election in a particular state and election 
year is won by the Democratic party.

\paragraph{Dataset.}
We obtained data describing the outcome of every United States 
Senate election from 1976 to 2016 (every two years, 
21 time periods) for all 50 
states \cite{US_senate_data}.
At most one Senator is elected per state per election,
and approximately 2/3 of the states elect a new Senator each election.
We created a training dataset consisting of the outcomes of every
Senate election from 1976 to 2012, and a test dataset using 2014 and 
2016.

\paragraph{Data records.}
In this problem, there is no $x$.
The outcome is $y \in \{0,1\}$,
with $y=1$ meaning the candidate from the Democratic party won.
The stratification feature $z$ is a tuple consisting of the state
(there are 50) and election year (there are 21);
for example, $z=(\mathrm{GA}, 1994)$ corresponds to the 1994 election in Georgia.
Here the number of stratification features
$K = 50 \cdot 21 = 1050$.
There are 639 training records
and 68 test records.

\paragraph{Data model.}
Our model is a simple Bernoulli model of the 
probability that the election goes to a Democrat
(see \S\ref{ss:bernoulli}).
For each state and election year, we have a single Bernoulli parameter
that can be interpreted as the probability that state
will elect a candidate from the Democratic party.
To be sure that our model never assigns zero likelihood,
we let $r(\theta)=0$ and $\Theta=[\epsilon,1-\epsilon]$, where
$\epsilon$ is some small positive constant; we used $\epsilon=\num{1e-5}$.
Only approximately 2/3 of states hold Senate
elections each election year, 
so our model makes a prediction of how a Senate 
race might unfold in years when a particular states' seats are 
not up for re-election.

\paragraph{Regularization graph.}
We take the Cartesian product of two regularization graphs:
\begin{itemize}
    \item \emph{State location}.
    The regularization graph is defined by a graph that has an edge
    with edge weight $\gamma_{\mathrm{state}}$ between two states if they share a border.
    (To keep the graph connected, we assume that Alaska borders Washington
    and that Hawaii borders California. There are 109 edges in this graph.)
    \item \emph{Year}.
    The regularization graph is a path graph, with edge weight $\gamma_{\mathrm{year}}$.
    (There are 20 edges in this graph.)
\end{itemize}

\begin{figure}
  \begin{center}
    \includegraphics[width=\textwidth]{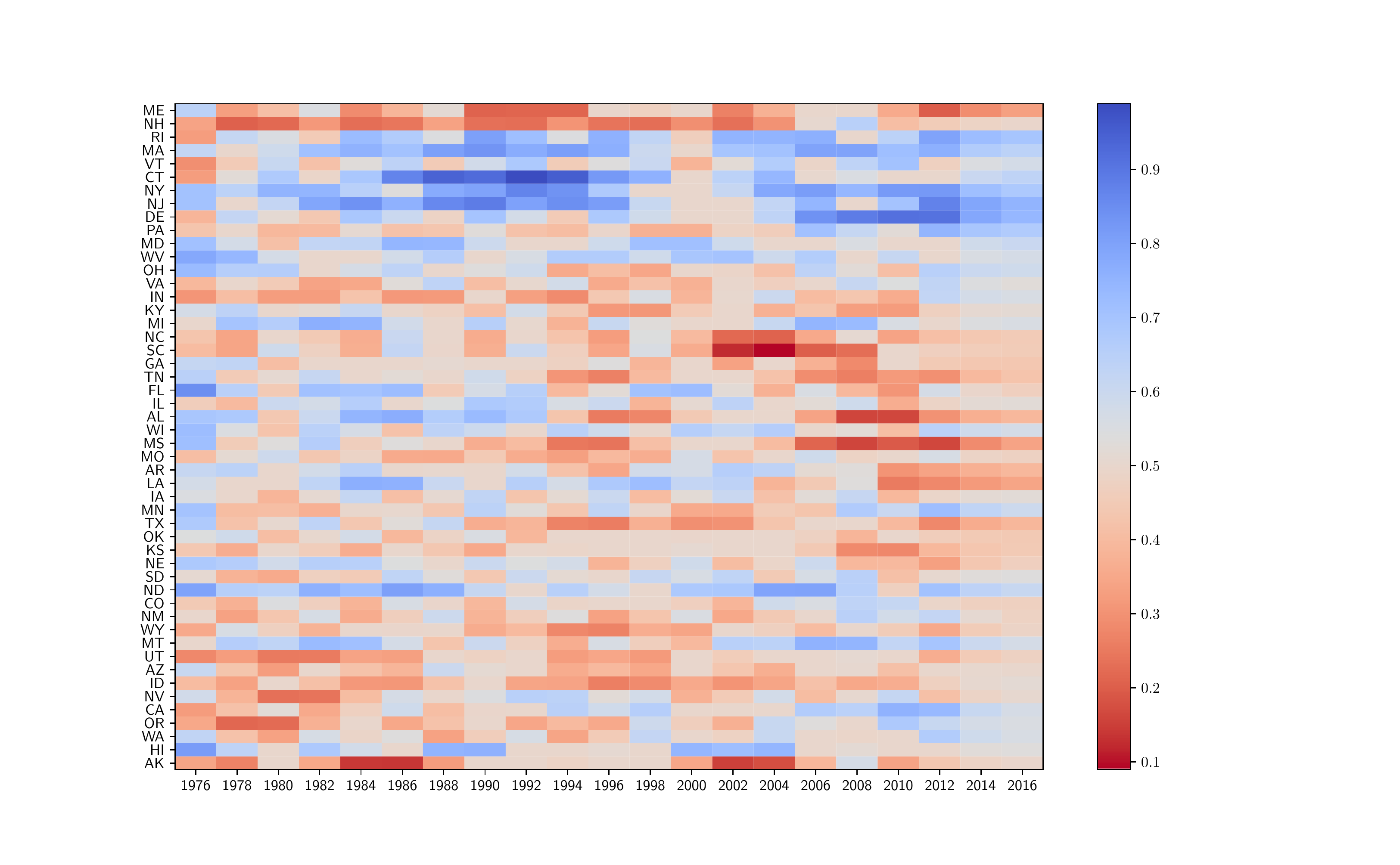}
  \caption{A heatmap representing the Bernoulli parameters across election 
  year and state.}
  \label{f-election_heatmap}
  \end{center}
\end{figure}

\paragraph{Results.}
We used $\gamma_{\mathrm{state}}=1$ and $\gamma_\mathrm{year}=4$.
Table~\ref{tab:elections} shows the train and test ANLL of the three models
on the training and test sets.
We see that the stratified model outperforms the other
two models.
Figure \ref{f-election_heatmap} shows a heatmap of the estimated Bernoulli
parameter in the stratified model for each state and election year.
The states are sorted according to the Fiedler eigenvector
(the eigenvector corresponding to the smallest nonzero eigenvalue)
of the Laplacian matrix of the state location regularization graph,
which groups states with similar parameters near each other.
High model parameters correspond to blue (Democrat) and low model parameters
correspond to red (Republican and other parties).
We additionally note that the model parameters for the entire test
years 2014 and 2016 were estimated using no data.

\begin{table}
\caption{Congressional elections results.}
  \vspace{.4em}
  \centering
  \begin{tabular}{lll}
    \toprule
    Model & Train ANLL & Test ANLL \\
    \midrule
    Separate  & 0.69 & 1.00 \\
    \textbf{Stratified} & \textbf{0.48} & \textbf{0.61} \\
    Common & 0.69 & 0.70 \\
    \bottomrule
  \end{tabular}
  \label{tab:elections}
\end{table}

\subsection{Chicago crime prediction}

We consider the problem of predicting the number of
crimes that will occur at a given location and time
in the greater Chicago area in the United States.

\paragraph{Dataset.}
We downloaded a dataset of crime records from the
greater Chicago area,
collected by the Chicago police department,
which include the time, location, and type of crime \cite{chicago2019crimes}.
There are 35 types of recorded crimes, ranging from theft and battery
to arson.
(For our purposes, we ignore the type of the crime.)
From the dataset, we created a training set, composed of the recorded
crimes in 2017, and a test set, composed of the recorded crimes in 2018.
The training set has \num{263752} records and the
test set has \num{262365} records.

\paragraph{Data records.}
We binned latitude/longitude into $20 \times 20$ equally sized bins.
Here there are no features $x$, the outcome $y$ is the number of crimes,
and the stratification feature $z$ is a tuple consisting of location bin,
week of the year, day of the week,
and hour of day.
For example, $z=((10, 10), 1, 2, 2)$ could correspond to a latitude between 41.796 and
41.811, a longitude between -87.68 and -87.667, on
the first week of January on a Tuesday between 1AM and 2AM.
Here the number of stratification features $K=20^2\cdot52\cdot7\cdot24=\num{3494400}$.

\paragraph{Creating the dataset.}
Recall that there are \num{263752} recorded crimes in our training data set.
However, the fact that there were no recorded
crimes in a time period is itself a data point.
We count the number of recorded crimes for
each location bin, day, and hour in 2017 (which could be zero),
and assign that number as the single data point for that stratification feature.
Therefore, we have \num{3494400} training data points,
one for each value of the stratification feature.
We do the same for the test set.

\paragraph{Data model.}
We model the distribution of the number of crimes $y$
using a Poisson distribution, as described
in \S\ref{sec:distribution_estimates}.
To be sure that our model never assigns zero log likelihood,
we let $r(\theta)=0$ and $\Theta = [\epsilon, \infty)$, where $\epsilon$ is some
small constant; we used $\epsilon=\num{1e-5}$.

\paragraph{Regularization graph.}
We take the Cartesian product of three regularization graphs:
\begin{itemize}
    \item \emph{Latitude/longitude bin}. The regularization
    graph is a two-dimensional grid graph, with edge weight $\gamma_\mathrm{loc}$.
    \item \emph{Week of the year}. The regularization graph is a
    cycle graph, with edge weight $\gamma_\mathrm{week}$.
    \item \emph{Day of the week}. The regularization graph is a
    cycle graph, with edge weight $\gamma_\mathrm{day}$.
    \item \emph{Hour of day}. The regularization graph is a cycle graph,
    with edge weight $\gamma_\mathrm{hour}$.
\end{itemize}
The Laplacian matrix has over 37 million nonzero entries
and the hyper-parameters are $\gamma_\mathrm{loc}$, $\gamma_\mathrm{week}$,
$\gamma_\mathrm{day}$, and $\gamma_\mathrm{hour}$.

\begin{table}
  \caption{Chicago crime results.}
  \vspace{.4em}
  \label{tab:crime}
  \centering
  \begin{tabular}{lll}
    \toprule
    Model & Train ANLL & Test ANLL \\
    \midrule
    Separate            & 0.068          & 0.740 \\
    \textbf{Stratified} & \textbf{0.221} & \textbf{0.234} \\
    Common              & 0.279          & 0.278 \\
    \bottomrule
  \end{tabular}
  \label{tab:crime_full}
\end{table}

\begin{figure}
  \begin{center}
    \includegraphics[]{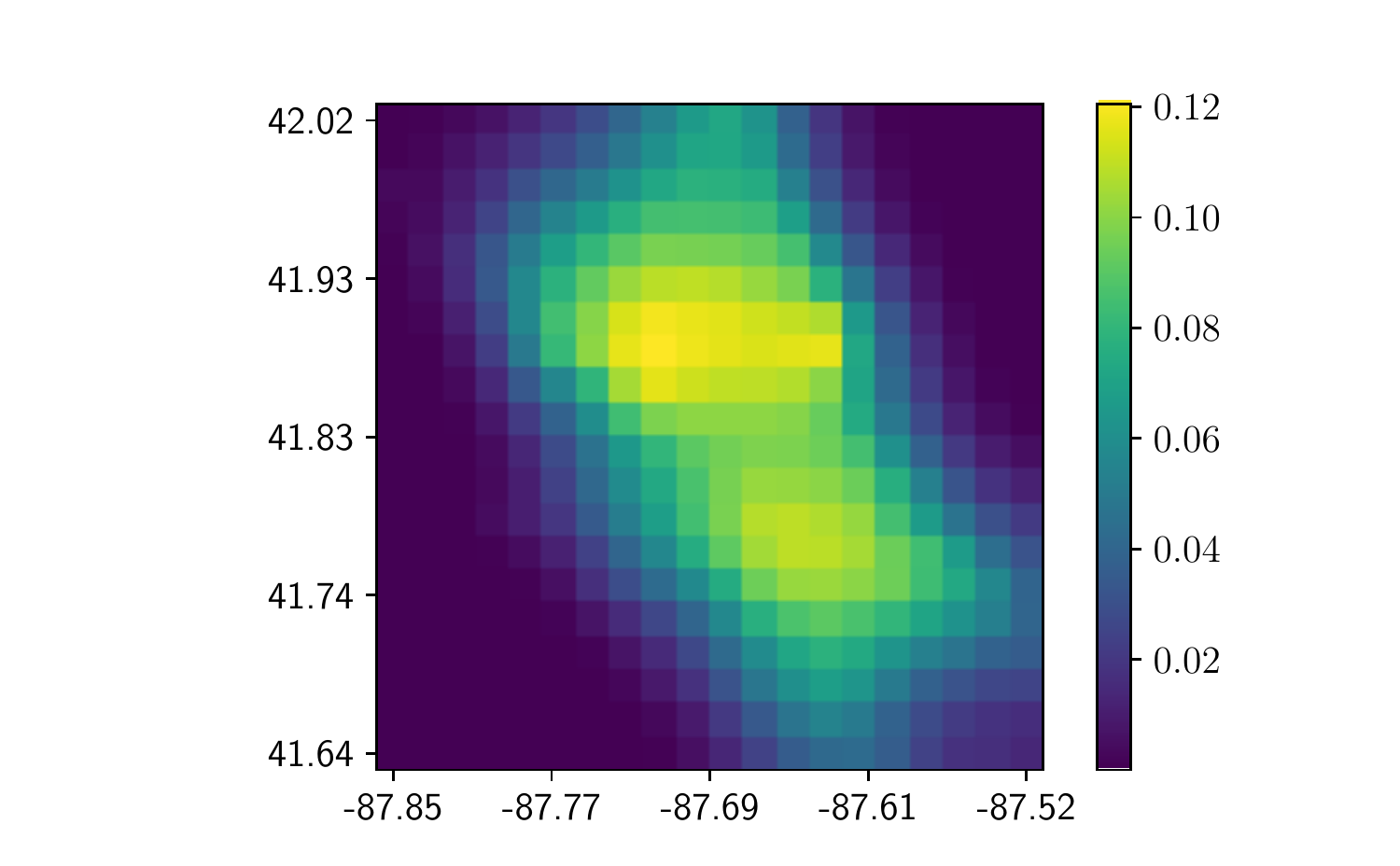}
  \caption{Rate of crime in Chicago for each latitude/longitude bin averaged
  over time, according to the stratified model.}
  \label{fig:crime_loc}
  \end{center}
\end{figure}

\begin{figure}
  \begin{center}
    \includegraphics[width=0.8\textwidth]{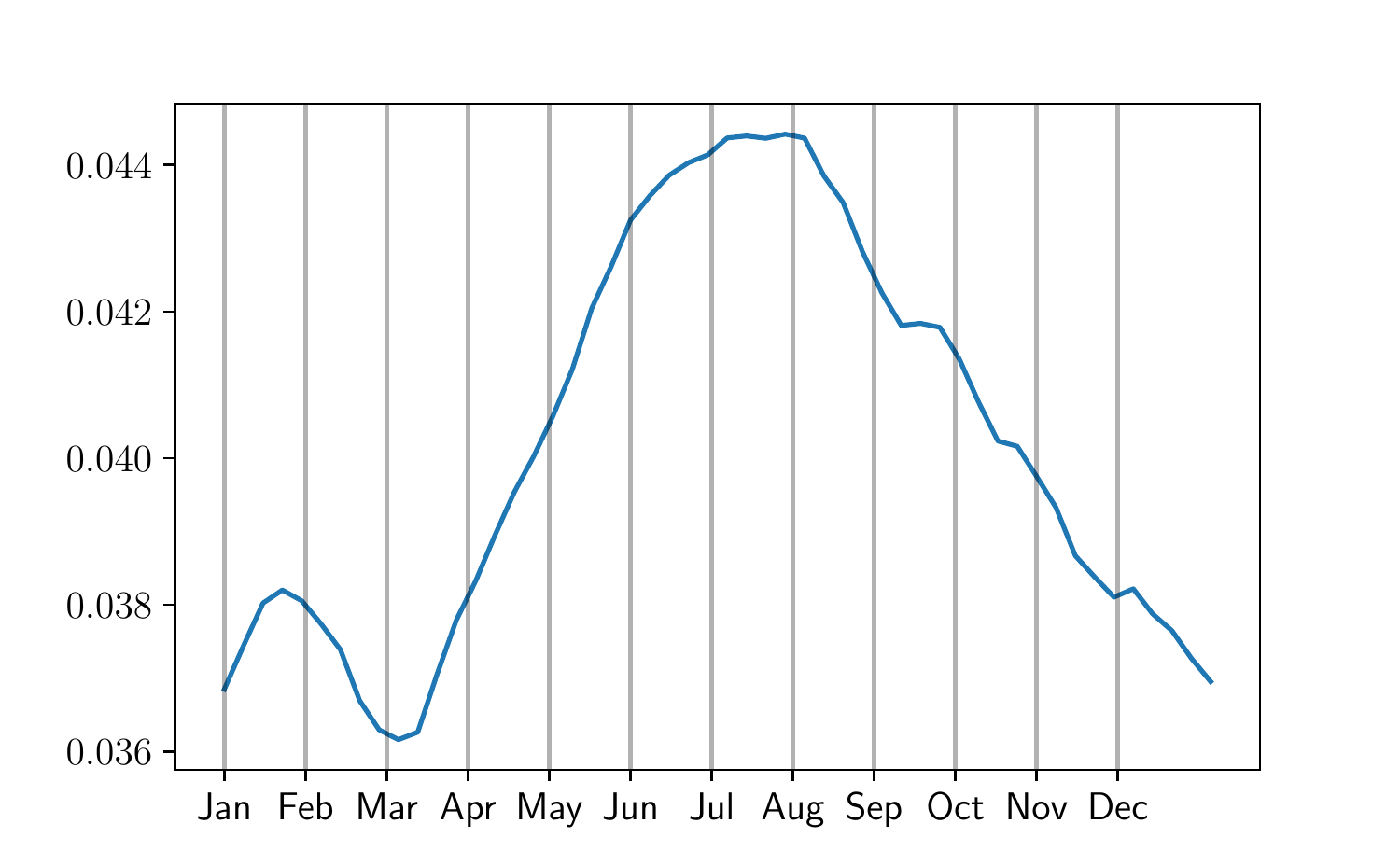}
  \caption{Rate of crime in Chicago versus week of the year averaged over
  latitude/longitude bins, according to the stratified model.}
  \label{fig:crime_week}
  \end{center}
\end{figure}

\begin{figure}
  \begin{center}
    \includegraphics[width=0.8\textwidth]{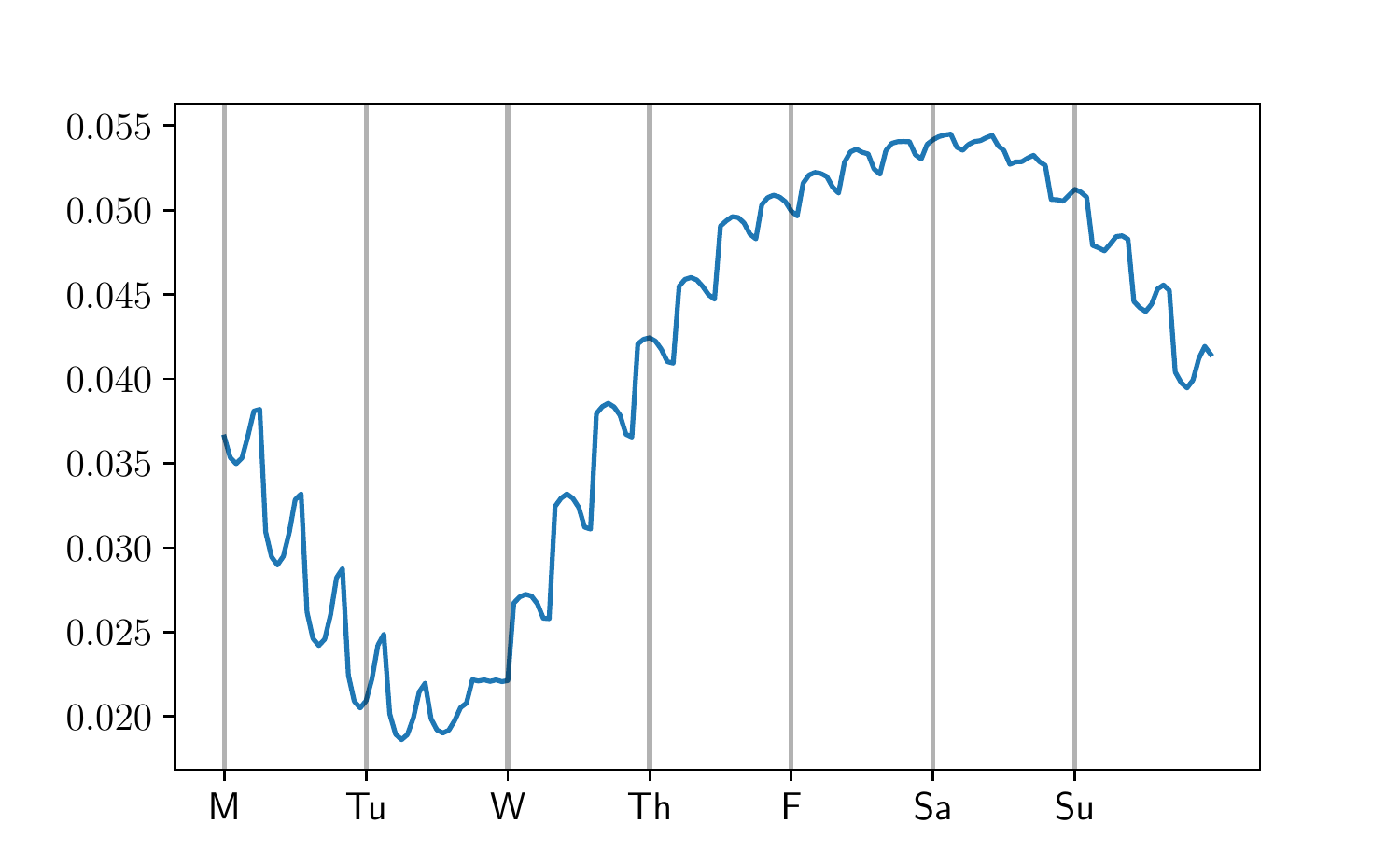}
  \caption{Rate of crime in Chicago versus hour of the week averaged over
  latitude/longitude bins, according to the stratified model.}
  \label{fig:crime_day}
  \end{center}
\end{figure}

\paragraph{Results.}
We ran the fitting method with
\[
\gamma_\mathrm{loc}=\gamma_\mathrm{week}=
\gamma_\mathrm{day}=\gamma_\mathrm{hour}=100.
\]
The method converged
to tolerances of $\epsilon_\mathrm{rel} = \num{1e-6}$
and $\epsilon_\mathrm{abs}=\num{1e-6}$ in 464 iterations,
which took about 434 seconds.
For each of the three models, we calculated the average negative log-likelihood (ANLL)
of the training and test set (see table \ref{tab:crime_full}).
We also provide visualizations of the
parameters of the fitted stratified model.
Figure \ref{fig:crime_loc} shows the rate of crime, according to each model,
in each location (averaged over time),
figure \ref{fig:crime_week} shows the rate of crime, according to each
model, for each week of the year (averaged over location and day),
and figure \ref{fig:crime_day} shows the rate of crime, according to each
model, for each hour of the day (averaged over location and week).
(This figure shows oscillations in crime rate with a period around 
8 hours.  We are not sure what this is, but suspect it may have
to do with work shifts.)

\section{Extensions and variations} \label{s-extensions}

In \S\ref{s-stratified_models} and \S\ref{s-distr-method}
we introduced the general idea of
stratified model fitting and a distributed fitting method.
In this section, we discuss possible
extensions and variations to the fitting
and the solution method.

\paragraph{Varied loss and regularizers.}
In our formulation of stratified model fitting, the local loss and regularizer
in the objective is a sum of the $K$ local losses and regularizers, which are 
all the same (save for problem data).
With slight alterations, the ideas described in this paper can extend to the case
where the model parameter for each stratification feature has different local loss
and regularization functions.

\paragraph{Nonconvex base fitting methods.}
We now explore the idea of fitting stratified models with
\emph{nonconvex} local loss and regularization
functions. The algorithm is exactly the same, except that we replace the
$\theta$ and $\tilde\theta$ update with approximate minimization.
ADMM no longer has the guarantees of
converging to an optimal point (or even converging at all),
so this method must be viewed as a
\emph{local} optimization method for stratified model fitting,
and its performance will depend heavily on the parameter initialization.
Despite the lack of convergence guarantees,
it can be very effective in practice, given a good approximate
minimization method.

\paragraph{Different sized parameter vectors.}
It is possible to
handle $\theta_k$ that have different sizes
with minor modifications to the ideas that we have described.
This can happen, \eg, when a feature is
not applicable to one category of the stratification.
A simple example is when one is stratifying on
sex and one of the features corresponds to number of pregnancies.
(It would not make sense to assign a number of pregnancies to men.)
In this case, the Laplacian regularization
can be modified so that only a subset of the entries of
the model parameters are compared.

\paragraph{Laplacian eigenvector expansions.}
If the Laplacian regularization term is small,
meaning that if $\theta_k$ varies smoothly over
the graph, then it can be well approximated by
a linear combination of a modest number, say $M$, of
the Laplacian eigenvectors.
The smallest eigenvalue is always zero, with
associated eigenvector $\ones$;
this corresponds to the common model, \ie, all $\theta_k$ are the same.
The other $M-1$ eigenvectors can be computed
from the graph, and we can change variables so that
the coefficients of the expansion are the variables.
This technique can potentially drastically reduce the number
of variables in the problem.

\paragraph{Coordinate descent Laplacian solver.}
It is possible to solve a regularized Laplacian system $Ax=b$
where $A=L+ (1/\lambda) I$ by applying randomized 
coordinate descent \cite{nesterov2012randomizedcoordinatedescent}
to the optimization problem
\[
\text{minimize} \; \; \; \frac{1}{2}x^TAx-b^Tx,
\]
which has solution $x=A^{-1}b$.
The algorithm picks a random row $i\in\{1,\ldots,K\}$,
then minimizes the objective over $x_i$, to give
\BEQ
x_i = \frac{b_i-\sum_{j\neq i} a_{ij} x_j}{a_{ii}},
\label{eq:update}
\EEQ
and repeats this process until convergence.
We observe that to compute the update for $x_i$, a vertex only needs the 
values of $x$ at its neighbors (since $a_{ij}$ is only nonzero for connected 
vertices).
This algorithm is guaranteed to converge under
very general conditions \cite{richtarik2014randomizedcoordinatedescent}.

\paragraph{Decentralized implementation.}

Algorithm \ref{a-dist_strat} can be implemented in a decentralized manner,
\ie, where each vertex only has access to its own parameters, dual variables,
and edge weights, and can only communicate these quantities to adjacent
vertices in the graph.
Line 1, line 2, and line 4 can be obviously be performed in parallel at
each vertex of the graph, using only local information.
Line 3 requires coordination across the graph, and can be implemented using a
decentralized version of the coordinate descent Laplacian solver described above.

The full decentralized implementation of the coordinate descent method goes as follows.
Each vertex broadcasts its value to adjacent vertices, and starts a random length timer.
Once a vertex's timer ends, it stops broadcasting and collects the values of the
adjacent vertices  (if one of them is not broadcasting, then it restarts the timer),
then computes the update \eqref{eq:update}.
It then starts another random length timer and begins broadcasting again.
The vertices agree beforehand on a periodic interval (\eg, every 15 minutes)
to complete line 3 and continue with the algorithm.

\section*{Acknowledgments} 
Shane Barratt is supported by the National Science Foundation Graduate 
Research Fellowship under Grant No. DGE-1656518.
The authors thank Trevor Hastie and Robert Tibshirani for helpful comments on an 
early draft of this paper.

\bibliography{strat_models}

\begin{thebibliography}{10}

\bibitem{legendre1805nouvelles}
A.~Legendre.
\newblock {\em Nouvelles m{\'e}thodes pour la d{\'e}termination des orbites des
  com{\`e}tes}.
\newblock 1805.

\bibitem{gauss1809theoria}
C.~Gauss.
\newblock {\em Theoria motus corporum coelestium in sectionibus conicis solem
  ambientium}, volume~7.
\newblock Perthes et Besser, 1809.

\bibitem{tibshirani1996lasso}
R.~Tibshirani.
\newblock Regression shrinkage and selection via the lasso.
\newblock {\em Journal of the Royal Statistical Society}, 58(1):267--288, 1996.

\bibitem{cox1958logistic}
D.~R. Cox.
\newblock The regression analysis of binary sequences.
\newblock {\em Journal of the Royal Statistical Society}, 20(2):215--242, 1958.

\bibitem{boser1992training}
B.~Boser, I.~Guyon, and V.~Vapnik.
\newblock A training algorithm for optimal margin classifiers.
\newblock In {\em Proceedings of the workshop on computational learning
  theory}, pages 144--152. ACM, 1992.

\bibitem{hastie2009elements}
T.~Hastie, R.~Tibshirani, and J.~Friedman.
\newblock Elements of statistical learning, 2009.

\bibitem{breiman1984classification}
L.~Breiman, J.~Friedman, C.J. Stone, and R.A. Olshen.
\newblock {\em Classification and Regression Trees}.
\newblock The Wadsworth and Brooks-Cole statistics-probability series. Taylor
  \& Francis, 1984.

\bibitem{goodfellow2016deeplearningbook}
I.~Goodfellow, Y.~Bengio, and A.~Courville.
\newblock {\em Deep Learning}.
\newblock MIT Press, 2016.

\bibitem{tobler1970computer}
W.~Tobler.
\newblock A computer movie simulating urban growth in the {D}etroit region.
\newblock {\em Economic geography}, 46(sup1):234--240, 1970.

\bibitem{boyd2004convex}
S.~Boyd and L.~Vandenberghe.
\newblock {\em Convex Optimization}.
\newblock Cambridge University Press, 2004.

\bibitem{boyd2011distributed}
S.~Boyd, N.~Parikh, E.~Chu, B.~Peleato, and J.~Eckstein.
\newblock Distributed optimization and statistical learning via the alternating
  direction method of multipliers.
\newblock {\em Foundation and Trends in Machine Learning}, 3(1):1--122, 2011.

\bibitem{kernan1999stratified}
W.~Kernan, C.~Viscoli, R.~Makuch, L.~Brass, and R.~Horwitz.
\newblock Stratified randomization for clinical trials.
\newblock {\em Journal of clinical epidemiology}, 52(1):19--26, 1999.

\bibitem{rothman1986epidemiology}
T.~L.~Lash K.~J.~Rothman, S.~Greenland.
\newblock {\em Modern Epidemiology}.
\newblock Lippincott Williams \& Wilkins, 3 edition, 1986.

\bibitem{kestenbaum2009epidemiology}
B.~Kestenbaum.
\newblock {\em Epidemiology and biostatistics: an introduction to clinical
  research}.
\newblock Springer Science \& Business Media, 2009.

\bibitem{jacob2007efficient}
L.~Jacob and J.~Vert.
\newblock Efficient peptide--{MHC-I} binding prediction for alleles with few
  known binders.
\newblock {\em Bioinformatics}, 24(3):358--366, 2007.

\bibitem{gross2016datasharedlasso}
S.~M. Gross and R.~Tibshirani.
\newblock Data shared lasso: A novel tool to discover uplift.
\newblock {\em Computational Statistics \& Data Analysis}, 101, 03 2016.

\bibitem{tibshirani2017pliablelasso}
R.~{Tibshirani} and J.~{Friedman}.
\newblock {A Pliable Lasso}.
\newblock {\em arXiv e-prints}, Jan 2018.

\bibitem{du2018pliablelasso}
W.~{Du} and R.~{Tibshirani}.
\newblock {A pliable lasso for the Cox model}.
\newblock {\em arXiv e-prints}, Jul 2018.

\bibitem{hallac2015network}
D.~Hallac, J.~Leskovec, and S.~Boyd.
\newblock Network lasso: Clustering and optimization in large graphs.
\newblock In {\em Proceedings of the ACM International Conference on Knowledge
  Discovery and Data Mining}, pages 387--396. ACM, 2015.

\bibitem{hallac2017network}
D.~Hallac, Y.~Park, S.~Boyd, and J.~Leskovec.
\newblock Network inference via the time-varying graphical lasso.
\newblock In {\em Proceedings of the ACM International Conference on Knowledge
  Discovery and Data Mining}, pages 205--213. ACM, 2017.

\bibitem{hallac2017snapvx}
D.~Hallac, C.~Wong, S.~Diamond, A.~Sharang, R.~Sosic, S.~Boyd, and J.~Leskovec.
\newblock {SnapVX}: A network-based convex optimization solver.
\newblock {\em Journal of Machine Learning Research}, 18(1):110--114, 2017.

\bibitem{leskovec2016snap}
J.~Leskovec and R.~Sosi{\v{c}}.
\newblock Snap: A general-purpose network analysis and graph-mining library.
\newblock {\em ACM Transactions on Intelligent Systems and Technology (TIST)},
  8(1):1, 2016.

\bibitem{diamond2016cvxpy}
S.~Diamond and S.~Boyd.
\newblock {CVXPY}: A {P}ython-embedded modeling language for convex
  optimization.
\newblock {\em Journal of Machine Learning Research}, 17(1):2909--2913, 2016.

\bibitem{hastie1993varyingcoefmodel}
T.~Hastie and R.~Tibshirani.
\newblock Varying-coefficient models.
\newblock {\em Journal of the Royal Statistical Society. Series B
  (Methodological)}, 55(4):757--796, 1993.

\bibitem{fan2008varyingcoefmodel}
J.~Fan and W.~Zhang.
\newblock Statistical methods with varying coefficient models.
\newblock {\em Statistics and Its Interface}, 1:179--195, 02 2008.

\bibitem{brunsdon1996geographically}
C.~Brunsdon, A.~Fotheringham, and M.~Charlton.
\newblock Geographically weighted regression: a method for exploring spatial
  nonstationarity.
\newblock {\em Geographical analysis}, 28(4):281--298, 1996.

\bibitem{mcmillen2004geographically}
D.~McMillen.
\newblock Geographically weighted regression: the analysis of spatially varying
  relationships, 2004.

\bibitem{YL:06}
M.~Yuan and Y.~Lin.
\newblock Model selection and estimation in regression with grouped variables.
\newblock {\em Journal of the Royal Statistical Society}, 68:49--67, 2006.

\bibitem{FHT:10}
J.~Friedman, T.~Hastie, and R.~Tibshirani.
\newblock {A note on the group lasso and a sparse group lasso}.
\newblock {\em arXiv e-prints}, 2010.

\bibitem{meier2008group}
L.~Meier, S.~Van De~Geer, and P.~B{\"u}hlmann.
\newblock The group lasso for logistic regression.
\newblock {\em Journal of the Royal Statistical Society}, 70(1):53--71, 2008.

\bibitem{belkin2006manifold}
M.~Belkin, P.~Niyogi, and V.~Sindhwani.
\newblock Manifold regularization: A geometric framework for learning from
  labeled and unlabeled examples.
\newblock {\em Journal of Machine Learning Research}, 7(Nov):2399--2434, 2006.

\bibitem{zhu2003semi}
X.~Zhu, Z.~Ghahramani, and J.~Lafferty.
\newblock Semi-supervised learning using {G}aussian fields and harmonic
  functions.
\newblock In {\em Proceedings of the International Conference on Machine
  Learning}, pages 912--919, 2003.

\bibitem{nadler2009semi_supervised_laplacian}
B.~Nadler, N.~Srebro, and X.~Zhou.
\newblock Statistical analysis of semi-supervised learning: The limit of
  infinite unlabelled data.
\newblock In Y.~Bengio, D.~Schuurmans, J.~D. Lafferty, C.~K.~I. Williams, and
  A.~Culotta, editors, {\em Advances in Neural Information Processing Systems},
  pages 1330--1338. Curran Associates, Inc., 2009.

\bibitem{boyd2006laplacianeigenvalues}
S.~Boyd.
\newblock Convex optimization of graph {L}aplacian eigenvalues.
\newblock In {\em Proceedings International Congress of Mathematicians}, pages
  1311--1319, 2006.

\bibitem{CWSS:11}
J.~Chen, C.~Wang, Y.~Sun, and X.~Shen.
\newblock Semi-supervised {L}aplacian regularized least squares algorithm for
  localization in wireless sensor networks.
\newblock {\em Computer Networks}, 55(10):2481 -- 2491, 2011.

\bibitem{ZF:15}
A.~Zouzias and N.~M. Freris.
\newblock Randomized gossip algorithms for solving {L}aplacian systems.
\newblock In {\em Proceedings of the European Control Conference}, pages
  1920--1925, 2015.

\bibitem{ZS:11}
D.~Zhang and D.~Shen.
\newblock Semi-supervised multimodal classification of alzheimer's disease.
\newblock In {\em 2011 IEEE International Symposium on Biomedical Imaging: From
  Nano to Macro}, pages 1628--1631, 2011.

\bibitem{WZSQS:12}
Z.~Wang, Z.~Zhou, X.~Sun, X.~Qian, and L.~Sun.
\newblock Enhanced lapsvm algorithm for face recognition.
\newblock {\em International Journal of Advancements in Computing Technology},
  4:343--351, 2012.

\bibitem{WSZQ:13}
Z.~Wang, X.~Sun, L.~Zhang, and X.~Qian.
\newblock Document classification based on optimal laprls.
\newblock {\em Journal of Software}, 8(4):1011--1018, 2013.

\bibitem{wang2011lapRLSmicrobio}
F.~Wang, Z.-A. Huang, X.~Chen, Z.~Zhu, Z.~Wen, J.~Zhao, and G.-Y. Yan.
\newblock {LRLSHMDA}: {L}aplacian regularized least squares for human
  microbe--disease association prediction.
\newblock {\em Scientific Reports}, 7(1):7601, 2017.

\bibitem{KS:11}
A.~Kovac and A.~D. A.~C. Smith.
\newblock Nonparametric regression on a graph.
\newblock {\em Journal of Computational and Graphical Statistics},
  20(2):432--447, 2011.

\bibitem{rudin1992nonlinear}
L.~Rudin, S.~Osher, and E.~Fatemi.
\newblock Nonlinear total variation based noise removal algorithms.
\newblock {\em Physica D: nonlinear phenomena}, 60(1-4):259--268, 1992.

\bibitem{Caruana1997}
R.~Caruana.
\newblock Multitask learning.
\newblock {\em Machine Learning}, 28(1):41--75, 1997.

\bibitem{ZY:17}
Y.~Zhang and Q.~Yang.
\newblock A survey on multi-task learning.
\newblock {\em Computing Research Repository}, abs/1707.08114, 2017.

\bibitem{sheldon2008graphical}
D.~Sheldon.
\newblock Graphical multi-task learning.
\newblock {\em Pre-print}, 2008.

\bibitem{spielman2007spectral}
D.~Spielman.
\newblock Spectral graph theory and its applications.
\newblock In {\em Foundations of Computer Science}, pages 29--38. IEEE, 2007.

\bibitem{teng2010laplacian}
S.~Teng.
\newblock The {L}aplacian paradigm: emerging algorithms for massive graphs.
\newblock In {\em International Conference on Theory and Applications of Models
  of Computation}, pages 2--14. Springer, 2010.

\bibitem{spielman2010algorithms}
D.~Spielman.
\newblock Algorithms, graph theory, and linear equations in {L}aplacian
  matrices.
\newblock In {\em Proceedings of the International Congress of Mathematicians},
  pages 2698--2722. World Scientific, 2010.

\bibitem{vishnoi2013lx}
N.~Vishnoi.
\newblock {$Lx=b$}: {L}aplacian solvers and their algorithmic applications.
\newblock {\em Foundations and Trends in Theoretical Computer Science},
  8(1--2):1--141, 2013.

\bibitem{CKMPP:14}
M.~B. Cohen, R.~Kyng, G.~L. Miller, J.~W. Pachocki, R.~Peng, A.~B. Rao, and
  S.~C. Xu.
\newblock Solving sdd linear systems in nearly {$m\log^{1/2}n$} time.
\newblock In {\em Proceedings of the Forty-sixth Annual ACM Symposium on Theory
  of Computing}, STOC '14, pages 343--352, New York, NY, USA, 2014. ACM.

\bibitem{STvDD:17}
M.~Schaub, M.~Trefois, P.~van Dooren, and J.~Delvenne.
\newblock Sparse matrix factorizations for fast linear solvers with application
  to {L}aplacian systems.
\newblock {\em SIAM Journal on Matrix Analysis and Applications},
  38(2):505--529, 2017.

\bibitem{sadhanala2016laplacian}
V.~Sadhanala, Y.-X. Wang, and R.~Tibshirani.
\newblock Graph sparsification approaches for {L}aplacian smoothing.
\newblock In Arthur Gretton and Christian~C. Robert, editors, {\em Proceedings
  of the 19th International Conference on Artificial Intelligence and
  Statistics}, volume~51 of {\em Proceedings of Machine Learning Research},
  pages 1250--1259, Cadiz, Spain, 09--11 May 2016. PMLR.

\bibitem{hestenes1952methods}
M.~Hestenes and E.~Stiefel.
\newblock Methods of conjugate gradients for solving linear systems.
\newblock {\em Journal of Research of the National Bureau of Standards}, 49(6),
  1952.

\bibitem{SBP:17}
Y.~Sun, P.~Babu, and D.~Palomar.
\newblock Majorization-minimization algorithms in signal processing,
  communications, and machine learning.
\newblock {\em IEEE Transactions in Signal Processing}, 65(3):794--816, 2017.

\bibitem{tuck2018distributed}
J.~{Tuck}, D.~{Hallac}, and S.~{Boyd}.
\newblock Distributed majorization-minimization for {L}aplacian regularized
  problems.
\newblock {\em IEEE/CAA Journal of Automatica Sinica}, 6(1):45--52, January
  2019.

\bibitem{LFYL:18}
C.~Lu, J.~Feng, S.~Yan, and Z.~Lin.
\newblock A unified alternating direction method of multipliers by majorization
  minimization.
\newblock {\em IEEE Transactions on Pattern Analysis and Machine Intelligence},
  40(3):527--541, 2018.

\bibitem{mm_tutorial:04}
D.~R. Hunter and K.~Lange.
\newblock A tutorial on mm algorithms.
\newblock {\em The American Statistician}, 58(1):30--37, 2004.

\bibitem{tikhonov1943inverse}
A.~N. Tikhonov.
\newblock On the stability of inverse problems.
\newblock {\em Doklady Akademii Nauk SSSR}, 39(5):195--198, 1943.

\bibitem{tikhonov1963regularization}
A.~N. Tikhonov.
\newblock Solution of incorrectly formulated problems and the regularization
  method.
\newblock {\em Soviet Mathematics Doklady}, 4:1035--1038, 1963.

\bibitem{boscovich1757litteraria}
R.~Boscovich.
\newblock De litteraria expeditione per pontificiam ditionem, et synopsis
  amplioris operis, ac habentur plura ejus ex exemplaria etiam sensorum
  impessa.
\newblock {\em Bononiensi Scientiarum et Artum Instuto Atque Academia
  Commentarii}, 4:353--396, 1757.

\bibitem{huber1964robust}
P.~Huber.
\newblock Robust estimation of a location parameter.
\newblock {\em The Annals of Mathematical Statistics}, 35(1):73--101, 1964.

\bibitem{vandenberghe1996semidefinite}
L.~Vandenberghe and S.~Boyd.
\newblock Semidefinite programming.
\newblock {\em SIAM review}, 38(1):49--95, 1996.

\bibitem{cortes1995svm}
C.~Cortes and V.~Vapnik.
\newblock Support-vector networks.
\newblock {\em Machine Learning}, 20(3):273--297, 1995.

\bibitem{hosmer2005logisticbook}
D.~W. Hosmer and S.~Lemeshow.
\newblock {\em Applied Logistic Regression}.
\newblock John Wiley and Sons, Ltd, 2005.

\bibitem{engel1988multinomiallogistic}
J.~Engel.
\newblock Polytomous logistic regression.
\newblock {\em Statistica Neerlandica}, 42(4):233--252, 1988.

\bibitem{Yang2007multiclassSVM}
X.-Y. Yang, J.~Liu, M.-Q. Zhang, and K.~Niu.
\newblock A new multi-class svm algorithm based on one-class svm.
\newblock In {\em Computational Science -- ICCS 2007}, pages 677--684, Berlin,
  Heidelberg, 2007. Springer Berlin Heidelberg.

\bibitem{manning2008multiclassSVM}
C.~D. Manning, P.~Raghavan, and H.~Sch\"{u}tze.
\newblock {\em Introduction to Information Retrieval}.
\newblock Cambridge University Press, 2008.

\bibitem{nelder1972generalized}
J.~Nelder and R.~Wedderburn.
\newblock Generalized linear models.
\newblock {\em Journal of the Royal Statistical Society}, 135(3):370--384,
  1972.

\bibitem{friedman2008sparse}
J.~Friedman, T.~Hastie, and R.~Tibshirani.
\newblock Sparse inverse covariance estimation with the graphical lasso.
\newblock {\em Biostatistics}, 9(3):432--441, 2008.

\bibitem{koopman1936exponentialfamily}
B.~O. Koopman.
\newblock On distributions admitting a sufficient statistic.
\newblock {\em Transactions of the American Mathematical Society},
  39(3):399--409, 1936.

\bibitem{pitman1936exponentialfamily}
E.~J.~G. Pitman.
\newblock Sufficient statistics and intrinsic accuracy.
\newblock {\em Mathematical Proceedings of the Cambridge Philosophical
  Society}, 32(4):567–579, 1936.

\bibitem{PB:14}
N.~Parikh and S.~Boyd.
\newblock Proximal algorithms.
\newblock {\em Foundations and Trends in Optimization}, 1(3):127--239, 2014.

\bibitem{HS:52}
M.~R. Hestenes and E.~Stiefel.
\newblock Methods of conjugate gradients for solving linear systems.
\newblock {\em Journal of Research of the National Bureau of Standards},
  49:409--436, 1952.

\bibitem{liu1989limited}
D.~Liu and J.~Nocedal.
\newblock On the limited memory bfgs method for large scale optimization.
\newblock {\em Mathematical programming}, 45(1-3):503--528, 1989.

\bibitem{he2000alternating}
B.~He, H.~Yang, and S.~Wang.
\newblock Alternating direction method with self-adaptive penalty parameters
  for monotone variational inequalities.
\newblock {\em Journal of Optimization Theory and applications},
  106(2):337--356, 2000.

\bibitem{wang2001decomposition}
S.~Wang and L.~Liao.
\newblock Decomposition method with a variable parameter for a class of
  monotone variational inequality problems.
\newblock {\em Journal of optimization theory and applications},
  109(2):415--429, 2001.

\bibitem{van2011numpy}
S.~Van Der~Walt, C.~Colbert, and G.~Varoquaux.
\newblock The numpy array: a structure for efficient numerical computation.
\newblock {\em Computing in Science \& Engineering}, 13(2):22, 2011.

\bibitem{jones2001scipy}
E.~Jones, T.~Oliphant, P.~Peterson, et~al.
\newblock {SciPy}: Open source scientific tools for {Python}, 2001.

\bibitem{hagberg2008exploring}
A.~Hagberg, P.~Swart, and D.~Chult.
\newblock Exploring network structure, dynamics, and function using {NetworkX}.
\newblock Technical report, Los Alamos National Lab.(LANL), Los Alamos, NM
  (United States), 2008.

\bibitem{paszke2017automatic}
A.~Paszke, S.~Gross, S.~Chintala, G.~Chanan, E.~Yang, Z.~DeVito, Z.~Lin,
  A.~Desmaison, L.~Antiga, and A.~Lerer.
\newblock Automatic differentiation in pytorch.
\newblock In {\em Advances in Neural Information Processing Systems}, 2017.

\bibitem{er2012mesothelioma}
O.~Er, A.~C. Tanrikulu, A.~Abakay, and F.~Temurtas.
\newblock An approach based on probabilistic neural network for diagnosis of
  mesothelioma's disease.
\newblock {\em Computers and Electrical Engineering}, 38(1):75 -- 81, 2012.
\newblock Special issue on New Trends in Signal Processing and Biomedical
  Engineering.

\bibitem{dua2019ucirepository}
D.~Dua and C.~Graff.
\newblock {UCI} machine learning repository.
\newblock \url{http://archive.ics.uci.edu/ml}, 2019.

\bibitem{US_senate_data}
{MIT}~Election Data and Science Lab.
\newblock {U.S. House} 1976–-2016, 2017.

\bibitem{chicago2019crimes}
Chicago~Police Department.
\newblock Crimes - 2001 to present.
\newblock
  \url{https://data.cityofchicago.org/Public-Safety/Crimes-2001-to-present/ijzp-q8t2},
  2019.
\newblock Accessed: 2019-04-09.

\bibitem{nesterov2012randomizedcoordinatedescent}
Y.~Nesterov.
\newblock Efficiency of coordinate descent methods on huge-scale optimization
  problems.
\newblock {\em SIAM Journal on Optimization}, 22(2):341--362, 2012.

\bibitem{richtarik2014randomizedcoordinatedescent}
P.~Richt{\'a}rik and M.~Tak{\'a}{\v{c}}.
\newblock Iteration complexity of randomized block-coordinate descent methods
  for minimizing a composite function.
\newblock {\em Mathematical Programming}, 144(1):1--38, 2014.

\bibitem{WT:09}
D.~M. Witten and R.~Tibshirani.
\newblock Covariance-regularized regression and classification for high
  dimensional problems.
\newblock {\em Journal of the Royal Statistical Society}, 71(3):615--636, 2009.

\end{thebibliography}

\newpage
\appendix
\section{Proximal operators}
\label{app-prox}

Recall that for $\theta \in \reals^n$, the proximal operator $\prox_{t f}: \reals^n \to \reals^n$ 
of a function $f: \reals^n \to \reals\cup\{+\infty\}$ and parameter $t > 0$ is defined as
\[
\prox_{t f}(\nu) = \underset{\theta \in \dom f}\argmin \left(t f(\theta) + (1/2)\|\theta-\nu\|_2^2 \right).
\]
In this section, we list some functions encountered in this paper and their associated proximal operators.

\paragraph{Indicator function of a convex set.} If $\mathcal C$ is a closed nonempty 
convex set, and we define $f(\theta)$ as 0 if $\theta \in \mathcal C$ and $+\infty$ otherwise, 
then
\[
\prox_{t f}(\nu) = \Pi_{\mathcal C}(\nu),
\]
\ie, the projection of $\nu$ onto $\mathcal C$.
(Note: If we allow $\mathcal C$ to be nonconvex, then $\prox_{t f}(\nu)$ is still of 
this form, but the projection is not guaranteed to be unique.)
For example, if $C=\reals_+^n$, then $(\prox_{t f}(\nu))_i = \max\{\nu_i,0\}$.

\paragraph{Sum of squares.} When $f(\theta) = (1/2)\|\theta\|_2^2$, the proximal operator is 
given by
\[
\prox_{t f}(\nu) = \frac{\nu}{t + 1}.
\]

\paragraph{$\ell_1$ norm.} When $f(\theta) = \|\theta\|_1$, the proximal operator is given by 
soft-thresholding, or
\[
\prox_{t f}(\nu) = (\nu-t)_+ - (-\nu-t)_+.
\]

\paragraph{$\ell_2$ norm.} The proximal operator of $f(\theta) = \|\theta\|_2$ is
\[
\prox_{t f}(\nu) = \left(1-\frac{t}{\|\nu\|_2}\right)_+ \nu.
\]

\paragraph{Negative logarithm.} The proximal operator of $f(\theta) = -\log \theta$ is
\[
\prox_{t f}(\nu) = \frac{\nu + \sqrt{\nu^2 + 4t}}{2}.
\]

\paragraph{Sum of $\ell_1$ and $\ell_2$-squared.} The proximal operator of 
$f(\theta) = \|\theta\|_1 + (\gamma/2)\|\theta\|_2^2$ for $\gamma \geq 0$, is
\[
\prox_{t f}(\nu) = \frac{1}{1+\gamma t}\prox_{t \|\cdot\|_1}(\nu).
\]

\paragraph{Quadratic.}
If $f(\theta) = (1/2)\theta^T A \theta + b^T \theta + c$, 
with $A \in \symm_+^n$, $b \in \reals^n$, and $c \in \reals$, then
\[
\prox_{t f}(\nu) = (I + t A)^{-1}(\nu-t b).
\]
If, \eg, $f(\theta) = \frac{1}{2}\|A\theta-b\|_2^2$, 
then $\prox_{t f}(\nu) = (I + t A^T A)^{-1}(\nu-t A^T b)$.

\paragraph{Poisson negative log likelihood.}
Here $f(\theta) = \sum_{i=1}^N\theta - y_i \log \theta$,
where $y_i \in \mathbb{Z}_+$ and $\theta > 0$.
If we let $S=\sum_{i=1}^N y_i$, then
\[
\prox_{t f}(\nu) = \frac{\nu-t N + \sqrt{(t N - \nu)^2 + 4 t S}}{2}.
\]

\paragraph{Covariance maximum likelihood.} 
Here
\[
f(\theta) = \Tr(S\theta)-\log\det(\theta).
\]
We want to find 
$\prox_{t f}(V) \coloneqq 
\underset{\theta}\argmin\left(\Tr(S\theta) - \log\det(\theta) + (1/2t)\|\theta-V\|_F^2\right)$.
The first-order optimality condition (along with the implicit constraint that $\theta \succ 0$) 
is that the gradient should vanish, \ie,
\[
S - \theta^{-1} + (1/t)(\theta - V) = 0.
\]
Rewriting, this is
\[
(1/t)\theta - \theta^{-1} = (1/t)V - S,
\]
implying that $\theta$ and $S - (1/t)V$ share the same eigenvectors \cite{WT:09},
stored in $Q$.
Let $w_i$ denote the $i$th eigenvalue of $\theta$, and $s_i$ the $i$th 
eigenvalue of $S - (1/t)V$.
Then we have
\[
1/w_i + (1/t)w_i = s_i, \quad i = 1, \ldots, n.
\]
To solve for $w = (w_1, \ldots, w_n)$, we solve these $n$ 1-dimensional quadratic equations, 
with the $i$th equation solved by
\[
w_i = \frac{t s_i + \sqrt{t^2 s_i^2 + 4t}}{2}.
\]
The proximal operator is then
\[
\prox_{t f}(V) = Q \diag(w) Q^T.
\]

\paragraph{Bernoulli negative log-likelihood.}
Here
\[
f(\theta) = -(\ones^T y) \log(\theta) - (n - \ones^T y) \log(1-\theta),
\]
with $y \in \reals^n$ is problem data and $\theta \in [0,1]$.
We want to find 
\[
\prox_{t f}(\nu) \coloneqq \underset{\theta}\argmin
\left(-(\ones^T y) \log(\theta) - (n - \ones^T y) \log(1-\theta) + (1/2t)(\theta-\nu)^2 \right).
\]
By setting the derivative of the above argument with respect to $\theta$ to zero, we find that
\[
\theta^3 - \theta^2(1-\nu) - \theta(nt - \nu) + t \ones^T y = 0.
\]
Solving this cubic equation (with respect to $\theta$) for the solution that lies in $[0,1]$ yields the
result of the proximal operator.

\end{document}